\documentclass[11pt,draftcls,onecolumn]{IEEEtran}

\usepackage{amsfonts}
\usepackage{subfigure}
\usepackage{amsmath}
\usepackage{amssymb}
\usepackage{amsfonts}
\usepackage{epsfig}
\usepackage{mathrsfs}
\usepackage{graphicx}
\usepackage{mathrsfs}
\usepackage{amsmath}
\usepackage{amsfonts}
\usepackage{amssymb}
\usepackage{graphicx}
\usepackage{amsmath,epsfig}
\usepackage{dsfont}
\usepackage{algorithm}
\usepackage{caption}
\usepackage{algpseudocode}

\def\bD{\mbox{\boldmath $D$}}

\def\b0{\mbox{\boldmath $0$}}

\def\bb{\mbox{\boldmath $b$}}
\def\bc{\mbox{\boldmath $c$}}
\def\bd{\mbox{\boldmath $d$}}

\def\bg{\mbox{\boldmath $g$}}

\def\bu{\mbox{\boldmath $u$}}
\def\bw{\mbox{\boldmath $w$}}

\def\bv{\mbox{\boldmath $v$}}

\newenvironment{proof}[1][Proof]{\noindent \textbf{#1.} }{\qedsymbol}
\newcommand{\qedsymbol}{\hspace{\fill}\rule{1.5ex}{1.5ex}}

\title{Sparse Distributed Learning Based on Diffusion Adaptation}

\author{ Paolo~Di Lorenzo,~\IEEEmembership{Student Member,~IEEE}, and Ali H. Sayed,~\IEEEmembership{Fellow,~IEEE}\\

\thanks{ Copyright (c) 2012 IEEE. Personal use of this material is permitted. However,
permission to use this material for any other purposes must be obtained from the
IEEE by sending a request to pubs-permissions@ieee.org. Paolo Di Lorenzo is with the Department of Information, Electronics, and Telecommunications (DIET), Sapienza University of Rome, Via Eudossiana 18, 00184 Rome, Italy. Ali H. Sayed is with the Electrical Engineering Department, University of California, Los Angeles, CA 90095. e-mail: {\tt dilorenzo@infocom.uniroma1.it, sayed@ee.ucla.edu}.

This work has been supported by FREEDOM Project, Nr. ICT-248891. The work of A.H. Sayed was supported in part by NSF grants CCF-1011918 and CCF-0942936. Part of this work was presented at the 2012 IEEE International Conference on Acoustic, Speech and Signal Processing, Kyoto, March 2012 \cite{Dilorenzo-Barbarossa-Sayed}.}}

\begin{document}

\maketitle

\vspace{-1.8cm}
\begin{abstract}
This article proposes diffusion LMS strategies for distributed estimation over adaptive networks that are able to exploit sparsity in the underlying system model. The approach relies on convex regularization, common in compressive sensing, to enhance the detection of sparsity via a diffusive process over the network.
The resulting algorithms endow networks with learning abilities and allow them to learn the sparse structure from the incoming data in real-time, and also to track variations in the sparsity of the model.  We provide convergence and mean-square performance analysis of the proposed method and show under what conditions it outperforms the unregularized diffusion version. We also show how to adaptively select the regularization parameter. Simulation results illustrate the advantage of the proposed filters for sparse data recovery.
\end{abstract}


\section{Introduction}

We consider the problem of distributed mean-square-error estimation, where a set of nodes is required to collectively estimate some vector parameter of interest from noisy measurements by relying solely on in-network processing. We consider an ad-hoc network consisting of $N$ nodes that are distributed over some geographic region. At every time instant $i$, every node $k$ collects a scalar measurement $d_k(i)$ of some random process $\bd_k(i)$ and a $1\times M$ regression vector $u_{k,i}$ of some random process $\bu_{k,i}$ with covariance matrix $R_{u,k}=\mathbb{E} \bu_{k,i}^* \bu_{k,i}>0$. The objective is for the nodes in the network to use the collected data $\{d_k(i),u_{k,i}\}$ to estimate some $M\times 1$ parameter vector $w^o$ in a distributed manner.

There are a couple of distributed strategies that have been developed in the literature for such purposes. One typical strategy is the incremental approach \cite{Bertsekas}-\hspace{-.1mm}\cite{Li-Chambers-Lopes-Sayed}, where each node communicates only with one neighbor at a time over a cyclic path.
In the incremental strategy, information is processed in a cyclic manner across the nodes of the network until optimization is achieved. However, determining a cyclic path that covers all nodes is an NP-hard problem \cite{Karp} and, in addition, cyclic trajectories are prone to link and node failures. When any of the edges along the path fails, the sharing of data through the cycle is interrupted and the algorithm stops performing.
To address these difficulties, adaptive diffusion techniques were proposed and studied in \cite{Lopes_Sayed,Cattivelli_Sayed}. In diffusion implementations, the nodes exchange information locally and cooperate with each other without the need for a central processor. In this way, information is processed on the fly by all nodes and the data diffuse across the network by means of a real-time sharing mechanism. The resulting adaptive networks exploit the time and spatial-diversity of the data more fully, thus endowing networks with powerful learning and tracking abilities. In view of their robustness and adaptation properties, diffusion networks have been applied to model a variety of self-organized behavior encountered in nature, such as birds flying in formation \cite{Cattivelli-Sayed3}, fish foraging for food \cite{Tu-Sayed} or bacteria motility \cite{Chen-Zhao-Sayed}. Diffusion adaptation has also been used to solve dynamic resource allocation problems in cognitive radios \cite{Dilo-Barb-Sayed}, to perform robust system modeling  \cite{Chouvardas-Slavakis-Theodoridis}, and to implement distributed learning over mixture models in pattern recognition applications \cite{Towfic-Chen-Sayed}.

In many situations, the parameter of interest, $w^o$, is sparse, containing only a few relatively large coefficients among many negligible ones.
Any prior information about the sparsity of $w^o$ can be exploited to help improve the estimation performance, as demonstrated in many recent efforts in the area of compressive sensing (CS) \cite{Donoho}-\hspace{-.1mm}\cite{Baraniuk}. Nevertheless, most CS efforts so far have concentrated on batch recovery methods, where the estimation of the desired vector is achieved from a collection of a fixed number of measurements. In this paper, we are instead interested in adaptive (online) techniques that allow the recovery of the vector $w^o$ to be pursued {\em both} recursively and distributively as new data arrive at the nodes. More importantly, we are interested in schemes that allow the recovery process to track changes in the sparsity pattern of the vector over time. Such schemes are useful in several contexts such as in the analysis of prostate cancer data \cite{Tibshirani}, \cite{Mateos-Bazerque-Giannakis}, spectrum sensing in cognitive radio \cite{Bazerque-Giannakis}, and spectrum estimation in wireless sensor networks \cite{Schizas-Mateos-Giannakis}.

Some of the early works that mix adaptation with sparsity-aware constructions include methods that rely on the heuristic selection of active taps \cite{Kawamura-Hatori}-\hspace{-.1mm}\cite{Li-Gu-Tang}, and on sequential partial updating techniques \cite{Etter}-\hspace{-.1mm}\cite{Godavarti-Hero}; some other methods assign proportional step-sizes to different taps according to their magnitudes, such as the proportionate normalized LMS (PNLMS) algorithm and its variations \cite{Duttweiler}-\hspace{-.1mm}\cite{Benesty-Gay}. In subsequent studies, motivated by the LASSO technique \cite{Tibshirani} and by connections with compressive sensing \cite{Baraniuk,Candes-Wakin-Boyd}, several algorithms for sparse adaptive filtering have been proposed based on LMS \cite{Chen-Gu-Hero}-\hspace{-.1mm}\cite{Shi-Shi}, RLS \cite{Angelosante-Bazerque-Giannakis,Babadi-Kalouptisidis-Tarokh}, and projection-based methods \cite{Kopsinis-Slavakis-Theodoridis}-\cite{Murakami-Yamagishi-Yukawa-Yamada}. A couple of distributed algorithms implementing LASSO over ad-hoc networks have also been considered before, although their main purpose has been to use the network to solve a {\em batch} processing problem \cite{Mota-Xavier-Aguiar-Puschel,Mateos-Bazerque-Giannakis}. One basic idea in  all these previously developed sparsity-aware techniques is to introduce a convex penalty term into the cost function to favor sparsity.

However, these earlier contributions did not consider the design of both {\em adaptive} and {\em distributed} solutions that are able to exploit and track sparsity while at the same time processing data in real-time and in a fully decentralized manner. Doing so would endow networks with learning abilities and would allow them to learn the sparse structure from the incoming data recursively and, therefore, to {\em track} variations in the sparsity pattern of the underlying vector as well. Investigations on adaptive and distributed solutions appear in \cite{Dilorenzo-Barbarossa-Sayed},\cite{Chouvardas-Slavakis-Kopsinis-Theodoridis}, and \cite{Liu-Li-Zhang}. In \cite{Dilorenzo-Barbarossa-Sayed}, we employed diffusion techniques that are able to identify and track sparsity over networks in a distributed manner; the techniques relied on the use of convenient convex regularization terms. In the related work \cite{Chouvardas-Slavakis-Kopsinis-Theodoridis}, the authors employ projection techniques onto hyperslabs and weighted $\ell_1$-balls to develop a useful sparsity-aware algorithm for distributed learning over diffusion networks. In \cite{Liu-Li-Zhang}, the authors use the same formulation of \cite{Dilorenzo-Barbarossa-Sayed} and the techniques of \cite{Lopes_Sayed}-\hspace{-.1mm}\cite{Cattivelli_Sayed} to independently arrive at useful diffusion strategies except that they limit the function $f(\cdot)$ in (\ref{glob_cost_function}) to choices of the form $\|w\|_p$, for particular selections of $p$-vector norms; they also include the regularization factor into the combination step of their algorithm rather than in the adaptation step, as done further ahead in this work. The algorithms proposed here and in \cite{Dilorenzo-Barbarossa-Sayed} are more general in a couple of respects: they allow for broader choices of the regularization function $f(\cdot)$, they allow for sharing of both data and weight estimates among the nodes (and not only estimates) by allowing for the use of two sets of combinations weights $\{a_{l,k},c_{l,k}\}$ instead of only one set, and the resulting mean-square and stability analyses are more demanding due to these generalizations; see, e.g.,  Appendices A and B. We further use the results of the analysis to propose an adaptive method to adjust online the important regularization parameter $\gamma$ in (\ref{glob_cost_function}). This is an important step in order to endow the resulting diffusion strategies with full adaptation abilities: adaptation to nonstationarities in the data and to changes in the sparsity patterns of the data.

The approach we follow in this work is based on  developing diffusion strategies that are stable in the mean-square-error sense, with guaranteed performance bounds.  For this reason, a detailed mean-square-error analysis is carried out in order to examine the behavior of the algorithm in the presence of noisy measurements and random regression data. The analysis ends up suggesting a convenient method for adapting the regularization parameter in a distributed manner as well. Doing so, enhances the sparsity-awareness of the algorithm and adds another useful layer of adaptation to the operation of the network.
In summary, in this paper we extend our preliminary work in \cite{Dilorenzo-Barbarossa-Sayed} to develop adaptive networks running diffusion algorithms subject to constraints that enforce sparsity. We consider two convex regularization constraints. In one case, we consider an $\ell_1$-norm regularization term, which acts as a uniform zero-attractor. In another case, and in order to improve the estimation performance, we employ reweighted regularization to selectively promote sparsity on the zero elements of $w^o$, rather than uniformly on all the elements. We provide detailed convergence analysis of the proposed methods, giving a closed form expression for the bias on the estimate due to regularization. We carry out a mean-square-error analysis, showing the conditions under which the sparse diffusion filter outperforms its unregularized version in terms of steady-state performance. It turns out that, if the system model is sufficiently sparse, it is possible to tune a single parameter to outperform the standard diffusion algorithm. Then, on the basis of this result, we propose a method to adaptively choose the regularization parameter.
In this way, the network is able to learn the sparse structure from the incoming data recursively and to adjust its parameters for improved tracking.  The main contributions of this paper are therefore: (a) exploitation of sparsity for distributed estimation over adaptive networks; (b) derivation of the mean-square properties of the sparse diffusion adaptive filter; (c) and adaptation of the regularization parameter to enhance performance under sparsity.

The paper is organized as follows. In Section II we develop the sparse diffusion algorithm for distributed adaptive estimation. Section III provides performance analysis, which includes mean stability, mean-square performance and adaptation of the regularization parameter. Section IV provides simulation results in support of the theoretical analysis, and section V draws some conclusions.

\textbf{Notation:} we use bold face letters to denote random variables and normal font letters to denote their realizations. Matrices and vectors are respectively denoted by capital and small letters.

\section{Sparse Distributed Estimation Over Adaptive Networks}

We assume the data $\{\bd_k(i),\bu_{k,i}\}$ collected by the various nodes are related to an unknown sparse vector $w^o$ via the linear model:
\begin{eqnarray}\label{scalar_obs}
\bd_{k}(i)=\bu_{k,i}w^o+\bv_k(i)
\end{eqnarray}
where $\bv_k(i)$ is a zero mean random variable with variance $\sigma^2_{v,k}$, independent of $\bu_{l,j}$ for all $l$ and $j$, and independent of $\bv_l(j)$ for $l\neq k$ and $i\neq j$. Linear models of the form (\ref{scalar_obs}) arise frequently in applications and are able to represent many cases of interest. The cooperative sparse estimation problem is cast as the search for the optimal estimator that minimizes in a fully distributed manner
the following cost function:
\begin{eqnarray}\label{glob_cost_function}
J^{\rm glob}(w)=\sum_{k=1}^{N}\mathbb{E}|\bd_{k}(i)-\bu_{k,i}w|^2+\gamma f(w)
\end{eqnarray}
where $\mathbb{E}(\cdot)$ denotes the expectation operator, and $f(w)$ is a real-valued convex regularization function weighted by the parameter $\gamma>0$, which is used to enforce sparsity of the solution.
The optimization problem in (\ref{glob_cost_function}) may be solved in a centralized fashion. In this approach, the nodes send their data to a central processor, or fusion center, where all computations can be performed. Centralized implementations of this type require transmitting data back and forth between the nodes and the central processor, which translates into requirements of power and bandwidth resources.  Additionally, centralized solutions have a serious point of failure: if the central processor fails, then the network operation is adversely affected and operation comes to a halt. For these reasons, we are interested in distributed solutions, where each node communicates with its neighboring nodes, and processing is distributed among all nodes in the network. In this way, communications are localized, and even when individual nodes fail, the network can continue to operate.

\subsection{Adaptive Diffusion Strategy}
We follow the approach proposed in \cite{Cattivelli_Sayed, Chen-Sayed} to derive distributed strategies for the minimization of $J^{\rm glob}(w)$ in (\ref{glob_cost_function}). We start by introducing an $N\times N$ matrix $C$ with non-negative entries $\{c_{l,k}\}$ such that
\begin{equation}\label{combination_coefficients}
c_{l,k}>0 \quad \hbox{if} \quad l\in\mathcal{N}_k,  \quad\quad C \mathds{1}=\mathds{1},
\end{equation}
where $\mathds{1}$ denotes the $N\times1$ vector with unit entries and $\mathcal{N}_k$ denotes the neighborhood of node $k$. Each coefficient $c_{l,k}$ represents a weight value that node $k$ assigns to information arriving from its neighbor $l$. Of course, the coefficient $c_{l,k}$ is equal to zero when nodes $l$ and $k$ are not connected. Furthermore, each row of $C$ adds up to one so that the sum of all weights leaving each node $l$ should be one. Using the coefficients in (\ref{combination_coefficients}), the global cost function in (\ref{glob_cost_function}) can be expressed as:
\begin{eqnarray}\label{glob_cost_function2}
J^{\rm glob}(w)=J_k^{\rm loc}(w)+\sum_{l\neq k}J_l^{\rm loc}(w)+\gamma f(w)
\end{eqnarray}
where
\begin{eqnarray}\label{loc_cost_function}
J_k^{\rm loc}(w)\triangleq\sum_{l=1}^{N}c_{l,k}\mathbb{E}|\bd_{l}(i)-\bu_{l,i}w|^2
\end{eqnarray}
The function introduced in (\ref{loc_cost_function}) is a local (neighborhood) cost for node $k$; it involves a weighted combination of the costs of its neighbors without considering the sparsity constraint. Assuming the processes $\bd_k(i)$ and $\bu_{k,i}$ are jointly wide sense stationary, the minimization of the local cost function (\ref{loc_cost_function}) over $w$ leads to the optimal local solution:
\begin{eqnarray}\label{loc_solution}
w_k^{\rm loc}=\left(\sum_{l=1}^Nc_{l,k}R_{u,l}\right)^{-1}\left(\sum_{l=1}^Nc_{l,k}r_{du,l}\right)
\end{eqnarray}
where $R_{u,k}=\mathbb{E}\bu^*_{k,i}\bu_{k,i}$ is assumed positive-definite (i.e., $R_{u,k}>0$) and $r_{du,k}=\mathbb{E}\bd_k(i)\bu^*_{k,i}$, where the operator $^*$ denotes complex-conjugate transposition. Thus, the local estimate $w_k^{\rm loc}$ is based solely on local covariance data $\{R_{u,l},r_{du,l}\}_{l\in {\cal N}_k}$ from the neighborhood of node $k$. If we multiply both sides of (\ref{scalar_obs}) by $\bu_{k,i}^*$ and take expectations and then add over the neighborhood of node $k$, it is easy to verify that the estimate $w_k^{\rm loc}$ from (\ref{loc_solution}) agrees with the desired vector $w^o$.  Therefore, in principle, each node $k$ can estimate $w^o$ {\em if} it knows the moments $\{R_{u,l},r_{du,l}\}$. Often, in practice, these moments are not available and nodes only sense realizations of data arising from these statistical distributions. In that case, cooperation among the nodes can help them improve their estimates of $w^o$ from the data realizations. To motivate the cooperative procedure, we start by noting that
a completion of squares argument shows that (\ref{loc_cost_function}) can be rewritten in terms of $w_k^{\rm loc}$ as
\begin{eqnarray}\label{loc_cost_function2}
J_k^{\rm loc}(w)=\|w-w_k^{\rm loc}\|^2_{\Gamma_k}+\hbox{\rm mmse}
\end{eqnarray}
where mmse is a constant term that does not depend on $w$, the notation $\|a\|^2_\Sigma=a^*\Sigma a$, for any nonnegative definite matrix $\Sigma$, and
\begin{eqnarray}
\Gamma_k\triangleq\sum_{l=1}^Nc_{l,k}R_{u,l}.
\end{eqnarray}
Thus, using (\ref{glob_cost_function2}), (\ref{loc_cost_function}) and (\ref{loc_cost_function2}), and dropping the constant mmse terms, we can replace the original global cost (\ref{glob_cost_function}) with the equivalent cost:
\begin{eqnarray}\label{glob_cost_function3}
J^{\rm glob'}(w)=\sum_{l\in\mathcal{N}_k}c_{l,k}\mathbb{E}|\bd_{l}(i)-\bu_{l,i}w|^2+\sum_{l\neq k}\|w-w_l^{\rm loc}\|^2_{\Gamma_l}+\gamma f(w)
\end{eqnarray}
Expression (\ref{glob_cost_function3}) shows how the local cost $J^{\rm loc}_k(w)$ can be modified to approach the desired global cost; two correction terms appear on the right-hand side: the regularization term $\gamma f(w)$ and a sum involving the local estimates $\{w_{l}^{\rm loc}\}$. Node $k$ cannot minimize (\ref{glob_cost_function3}) directly. This is because  the cost in (\ref{glob_cost_function3}) still requires the nodes to have access to global information, namely, the local estimates $\{w_{l}^{\rm loc}\}$, and the matrices $\{\Gamma_l\}$, from all other nodes in the network. To enable a distributed and iterative procedure, we make three adjustments to (\ref{glob_cost_function3}).

First, we limit the sum over $l\neq k$ to a sum over the neighbors of node $k$, i.e., only over $l\in{\cal N}_k$. This step is reasonable since node $k$ can only rely on data that are available to it from its neighborhood. Second, we replace the covariance matrices $\Gamma_l$ with constant diagonal weighting matrices of the form $\Gamma_l=b_{l,k}I_M$, where $b_{l,k}$ is a set of non-negative real coefficients that give different weights to different neighbors, and $I_M$ is the $M\times M$ identity matrix. Although the $\{\Gamma_l\}$ from its neighbors are available to node $k$, this step is meant to simplify the structure of the resulting algorithm. This substitution is also reasonable in view of the fact that norms are equivalent and that each of the weighted norms in (9) can be bounded as
\begin{equation}
\lambda_{\min}(\Gamma_l)\cdot\|w-w_{l}^{\rm loc}\|^2\leq\|w-w_{l}^{\rm loc}\|^2_{\Gamma_l}\leq \lambda_{\max}(\Gamma_l)\cdot\|w-w_{l}^{\rm loc}\|^2
\end{equation}
Substitutions of this kind are common in the stochastic approximation literature where Hessian matrices, such as $\Gamma_l$, are replaced by multiples of the identity matrix; such approximations allow the use of simpler steepest-descent iterations in place of Newton-type iterations \cite{Chen-Sayed}. At this stage, we do not need to worry about the selection of the weights $\{b_{l,k}\}$ because they are going to be embedded into another set of coefficients that the designer can choose.
Finally, while the nodes are attempting to estimate $w^o$, they do not know what the optimal local estimates $w_{l}^{\rm loc}$ are during the iterative learning process. As the ensuing discussion will reveal,
each node in the resulting distributed algorithm will be working on estimating the sparse vector $w^o$ and will have access to a local estimate for $w^o$, which we denote by $\psi_l$ at node $l$. Due to the diffusion process, this estimate will not be based solely on data from the neighborhood of node $l$ but also on data from across the network.
We are therefore motivated to replace $w_l^{\rm loc}$ in (\ref{glob_cost_function3}) by $\psi_l$. In this way, each node $k$ can instead minimize the following modified local cost function:
\begin{eqnarray}\label{glob_cost_function4}
J^{\rm dist}_k(w)=\sum_{l\in\mathcal{N}_k}c_{l,k}\mathbb{E}|\bd_{l}(i)-\bu_{l,i}w|^2+\sum_{l\in\mathcal{N}_k/\{k\}}b_{l,k}\|w-\psi_l\|^2+\gamma f(w).
\end{eqnarray}
The cost in (\ref{glob_cost_function4}) is now defined in terms of information that is available to node $k$.
Observe that while (\ref{glob_cost_function4}) is a local approximation for the global cost (\ref{glob_cost_function3}), it is nevertheless more general than the local cost (\ref{loc_cost_function}). The node $k$ can then proceed to optimize (\ref{glob_cost_function4}) by means of a steepest-descent procedure. Note that all functions in (\ref{glob_cost_function4}) are continuously differentiable except possibly $f(w)$, which is only supposed to be convex. Thus, computing the sub-gradient of (\ref{glob_cost_function4}) we obtain
\begin{eqnarray}\label{subgradient}
[\nabla J^{\rm dist}_k(w)]^*=\sum_{l\in\mathcal{N}_k}c_{l,k}(R_{u,l}w-r_{du,l})+\sum_{l\in\mathcal{N}_k/\{k\}}b_{l,k}(w-\psi_l)+\gamma \partial f(w)
\end{eqnarray}
where $\partial f(w)$ is the sub-gradient of the convex function $f(w)$. Then, we can use (\ref{subgradient}) to obtain a steepest descent recursion for the estimate of $w^o$ at node $k$ at time $i$, denoted by $w_{k,i}$, such as
\begin{eqnarray}\label{steepest_descent}
w_{k,i}=w_{k,i-1}+\mu_k \displaystyle \sum_{l\in\mathcal{N}_k}c_{l,k}(r_{du,l}-R_{u,l}w_{k,i-1})+\mu_k\sum_{l\in\mathcal{N}_k/\{k\}}b_{l,k}(\psi_l-w_{k,i-1})-\mu_k\gamma \partial f(w_{k,i-1})
\end{eqnarray}
for some sufficiently small positive step-sizes $\{\mu_k\}$. The update (\ref{steepest_descent}) involves the sum of three terms and we can compute it in two steps by generating an intermediate estimate $\psi_{k,i}$, as follows:
\begin{eqnarray}
\psi_{k,i}&=&w_{k,i-1}+\mu_k \displaystyle \sum_{l\in \mathcal{N}_k}c_{l,k} (r_{du,l}-R_{u,l}w_{k,i-1})-\mu_k\gamma \partial f(w_{k,i-1})\label{steepest_descent2_1} \\
w_{k,i}&=&\psi_{k,i}+\mu_k\displaystyle\sum_{l \in {\cal N}_k/\{k\}}b_{l,k}(\psi_l-w_{k,i-1}) \label{steepest_descent2_2}
\end{eqnarray}
Since every node in the network will be running recursions of the form (\ref{steepest_descent2_1})-(\ref{steepest_descent2_2}), then the intermediate estimate of $w^o$ that is available to each node $l$ at time $i$ is $\psi_{l,i}$. Therefore, we replace $\psi_l$ in (\ref{steepest_descent2_2}) by $\psi_{l,i}$.
Moreover, since $\psi_{k,i}$ is an updated estimate relative to $w_{k,i-1}$, as evidenced by (\ref{steepest_descent2_1}), we are motivated to replace
$w_{k,i-1}$ in (\ref{steepest_descent2_2}) by $\psi_{k,i}$, which generally leads to enhanced performance since $\psi_{k,i}$ contains more information than $w_{k,i-1}$. This step is reminiscent of an incremental-type substitution \cite{Bertsekas}-\hspace{-.1mm}\cite{Li-Chambers-Lopes-Sayed}. Performing these substitutions in (\ref{steepest_descent2_2}), we get:
\begin{eqnarray}
w_{k,i}=\psi_{k,i}+\mu_k\displaystyle\sum_{l \in {\cal N}_k/\{k\}}b_{l,k}(\psi_{l,i}-\psi_{k,i})
\end{eqnarray}
If we now introduce the entries of an $N\times N$ matrix $A=\{a_{l,k}\}$ such that
\begin{eqnarray}
a_{l,k}\triangleq\mu_k b_{l,k} \quad (l\neq k), \quad\quad  a_{k,k}\triangleq 1-\mu_k \sum_{l\in {\cal N}_k} b_{l,k},\;\;\;\;\;\;
a_{l,k}=0\;\;\mbox{\rm if}\;\;l\notin{\cal N}_k
\end{eqnarray}
then, we can rewrite the update in (\ref{steepest_descent2_1})-(\ref{steepest_descent2_2}) as:
\begin{eqnarray} \label{steepest_descent3}
\begin{cases}
\psi_{k,i}=w_{k,i-1}+\mu_k \displaystyle \sum_{l\in \mathcal{N}_k}c_{l,k} (r_{du,l}-R_{u,l}w_{k,i-1})-\mu_k\gamma \partial f(w_{k,i-1}) \\
w_{k,i}= \displaystyle\sum_{l \in {\cal N}_k}a_{l,k}\psi_{l,i}
\end{cases}
\end{eqnarray}
where the weighting coefficients $\{a_{l,k},c_{l,k}\}$ are real, non-negative and satisfy:
\begin{equation}\label{combination_coefficients2}
c_{l,k}>0,\quad a_{l,k}>0  \quad \hbox{if} \quad l\in\mathcal{N}_k, \quad C \mathds{1}=\mathds{1}, \quad A^T\mathds{1}=\mathds{1}.
\end{equation}
The recursion in (\ref{steepest_descent3}) requires knowledge of the second-order moments $\{R_{u,k},r_{du,k}\}$. An adaptive implementation can be obtained by replacing these second-order moments by local instantaneous approximations, say, of the LMS type, as follows:
\begin{eqnarray}\label{inst_approx}
R_{u,k}\simeq u^*_{k,i}u_{k,i}, \quad\quad r_{du,k}\simeq d_k(i)u^*_{k,i}.
\end{eqnarray}
Thus, substituting the approximations (\ref{inst_approx}) into (\ref{steepest_descent3}), we arrive at the following Adapt-then-Combine (ATC) strategy. We refer to the algorithm as the ATC sparse diffusion algorithm.
\begin{algorithm}
\caption*{\textbf{ATC sparse diffusion LMS}}
\vspace{.3cm}
Start with $w_{k,-1}=0$ for all $k$. Given non-negative real coefficients $\{a_{l,k},c_{l,k}\}$ satisfying (\ref{combination_coefficients2}), for each time $i\geq0$ and for each node $k$, repeat:
\begin{eqnarray}\label{ATC diffusion}
\begin{cases}
\psi_{k,i}=w_{k,i-1}+\mu_k \displaystyle \sum_{l\in \mathcal{N}_k}c_{l,k} u^*_{l,i}[d_{l}(i)-u_{l,i}w_{k,i-1}]-\mu_k\gamma \partial f(w_{k,i-1}) \hspace{1cm} \hbox{(adaptation step)} \\
w_{k,i}=\displaystyle\sum_{l \in {\cal N}_k}a_{l,k}\psi_{l,i} \hspace{8.85cm} \hbox{(diffusion step)}
\end{cases}
\end{eqnarray}
\end{algorithm}
The first step in (\ref{ATC diffusion}) is an adaptation step, where the coefficients $c_{l,k}$ determine which nodes $l\in \mathcal{N}_k$ should share their measurements $\{d_{l}(i),u_{l,i}\}$ with node $k$. The second step is a diffusion step where the intermediate estimates $\psi_{l,i}$, from the neighbors $l\in \mathcal{N}_k$, are combined through the coefficients $\{a_{l,k}\}$. We remark that had we reversed the steps (\ref{steepest_descent2_1}) and (\ref{steepest_descent2_2}) to implement (\ref{steepest_descent}), we arrive at a similar but alternative strategy, known as the Combine-then-Adapt (CTA) strategy; in this implementation, the only difference is that data aggregation is performed before adaptation (see, e.g., \cite{Cattivelli_Sayed}).
\begin{algorithm}
\caption*{\textbf{CTA sparse diffusion LMS}}
\vspace{.3cm}
Start with $w_{k,-1}=0$ for all $k$. Given non-negative real coefficients $\{a_{l,k},c_{l,k}\}$ satisfying (\ref{combination_coefficients2}), for each time $i\geq0$ and for each node $k$, repeat:
\begin{eqnarray}\label{CTA diffusion}
\begin{cases}
\psi_{k,i-1}=\displaystyle\sum_{l \in {\cal N}_k}a_{l,k}w_{l,i-1} \hspace{8.2cm} \hbox{(diffusion step)} \\
w_{k,i}=\psi_{k,i-1}+\mu_k \displaystyle \sum_{l\in \mathcal{N}_k}c_{l,k} u^*_{l,i}[d_{l}(i)-u_{l,i}\psi_{k,i-1}]-\mu_k\rho \partial f(\psi_{k,i-1}) \hspace{1.2cm} \hbox{(adaptation step)}
\end{cases}
\end{eqnarray}
\end{algorithm}
The complexity of the sparse diffusion schemes in (\ref{ATC diffusion})-(\ref{CTA diffusion}) is $O(3M)$, which is the same complexity as standard stand-alone LMS adaptation.
It was argued in \cite{Cattivelli_Sayed} that ATC strategies generally outperform CTA strategies. For this reason, we continue our discussion by focusing on the ATC algorithm (\ref{ATC diffusion}); similar analysis applies to CTA.

Compared with the strategies proposed in \cite{Chouvardas-Slavakis-Kopsinis-Theodoridis} and \cite{Liu-Li-Zhang}, the diffusion algorithm (\ref{ATC diffusion}) exploits data in the neighborhood more fully; the adaptation step aggregates data $\{d_l(i),u_{l,i}\}$ from the neighbors, and the diffusion step aggregates estimates $\{\psi_{l,i}\}$ from the same neighbors. The implementation in \cite{Chouvardas-Slavakis-Kopsinis-Theodoridis} uses a different algorithmic structure with $C=I$ so that data $\{d_l(i),u_{l,i}\}$ from the neighbors are not directly used. Compared with \cite{Liu-Li-Zhang}, observe that the effect of the regularization factor in (\ref{ATC diffusion}) influences the adaptation step, and not the combination step as in \cite{Liu-Li-Zhang}. Observe also that the adaptation step allows for the exchange of data $\{d_{l}(i),u_{l,i}\}$ among the nodes through the use of the coefficients $\{c_{l,k}\}$, whereas \cite{Liu-Li-Zhang} uses $C=I$ as well.

\subsection{Sparse Regularization}

Before proceeding with the discussions, let us comment on the regularization function $f(w)$ in (\ref{glob_cost_function}). A sparse vector $w^o$ generally contains only a few relatively large coefficients interspersed among many negligible ones and the location of the non-zero elements is often unknown beforehand. However, in some applications, we may have available some upper bound on the number of nonzero elements. Thus, assume that $w^o$ satisfies
\begin{equation}\label{l0norm}
\|w^o\|_0\leq \tau,
\end{equation}
where $\|\cdot\|_0$ is the $\ell_0$-norm, denoting the number of non-zero entries of a vector, and $\tau$ is a known upper bound. Since the $\ell_0$-norm in (\ref{l0norm})
is not convex, we cannot use it directly. Thus, motivated by LASSO \cite{Tibshirani} and work on compressive sensing \cite{Baraniuk}, we first consider the following $\ell_1$-norm convex choice for a regularization function:
\begin{equation}\label{l1norm}
f_1(w)=\|w\|_1\triangleq\sum_{m=1}^{M}|w_m|
\end{equation}
which amounts to the sum of the absolute entries of the vectors. The $\ell_1$-norm works as a surrogate approximation for the $\ell_0$-norm. This choice leads to an algorithm update in (\ref{ATC diffusion}) where the subgradient column vector is given by
\begin{equation}\label{sign}
\partial f_1(w)={\rm sign}(w)
\end{equation}
and the entries of the vector $\mbox{\rm sign}(w)$ are obtained by applying the following function to each entry of $w$:
\begin{eqnarray}\label{sign2}
{\rm sign}(w_m)=\begin{cases}
w_m/|w_m|, \quad w_m\neq0\\
0,  \quad\quad\hspace{.25cm} w_m=0
\end{cases}
\end{eqnarray}
This update leads to what we shall refer to as the \textit{zero-attracting} (ZA) diffusion algorithm. The ZA update uniformly shrinks all components of the vector, and does not distinguish between zero and non-zero elements \cite{Candes-Wakin-Boyd, Chen-Gu-Hero}. Since all the elements are forced toward zero uniformly, the performance would deteriorate for systems that are not sufficiently sparse. Motivated by the idea of reweighting in compressive sampling \cite{Candes-Wakin-Boyd, Chen-Gu-Hero},\cite{Kopsinis-Slavakis-Theodoridis}, we also consider the following  approximation:
\begin{equation}\label{approx_l0norm}
\|w\|_0\simeq\sum_{m=1}^M \frac{|w_m|}{\varepsilon+|w_m|}
\end{equation}
which, for very small positive values of $\varepsilon$, is a better approximation for the $\ell_0$-norm of a vector $w$ than the $\ell_1$-norm \cite{Candes-Wakin-Boyd}, thus enhancing sparse recovery by the algorithm. Therefore, interpreting (\ref{approx_l0norm}) as a weighted $\ell_1$-norm regularization, to update the algorithm in (\ref{ATC diffusion}), we shall consider the following sub-gradient column vector:
\begin{equation}\label{update_rew_l1norm}
\partial f_2(w)=\mbox{\rm diag}\left\{\frac{1}{\varepsilon+|w_1|},
\frac{1}{\varepsilon+|w_2|},
\ldots,
\frac{1}{\varepsilon+|w_M|},
\right\}\cdot
\mbox{\rm sign}(w)
\end{equation}
This choice leads to what we shall refer to as the \textit{reweighted zero-attracting} (RZA) diffusion algorithm. The update in (\ref{update_rew_l1norm}) selectively shrinks only the components whose magnitudes are comparable to $\varepsilon$, and there is little effect on components satisfying $|w_m|\gg \varepsilon$, see, e.g., \cite{Candes-Wakin-Boyd, Chen-Gu-Hero, Kopsinis-Slavakis-Theodoridis, Dilorenzo-Barbarossa-Sayed, Liu-Li-Zhang}.

\section{Mean-Square Performance Analysis}

From now on, we view the estimates $w_{k,i}$ as realizations of a random process $\bw_{k,i}$ and analyze the performance of the sparse diffusion algorithm in terms of its mean-square behavior.
To do so, we introduce the error quantities $\tilde{\bw}_{k,i}=w^o-\bw_{k,i}$, $\tilde{\boldsymbol{\psi}}_{k,i}=w^o-\boldsymbol{\psi}_{k,i}$, and the network vectors:
\begin{eqnarray}\label{err_vectors}
\bw_{i}=\begin{bmatrix} \bw_{1,i} \\ \vdots \\ \bw_{N,i}  \end{bmatrix},\hspace{.4cm}
\tilde{\bw}_{i}=\begin{bmatrix} \tilde{\bw}_{1,i} \\ \vdots \\ \tilde{\bw}_{N,i}  \end{bmatrix},\hspace{.4cm}
\tilde{\boldsymbol{\psi}}_{i}=\begin{bmatrix} \tilde{\boldsymbol{\psi}}_{1,i} \\ \vdots \\ \tilde{\boldsymbol{\psi}}_{N,i}  \end{bmatrix}
\end{eqnarray}
We also introduce the block diagonal matrix
\begin{eqnarray}\label{step_matrix}
\mathcal{M}={\rm diag}\{\mu_1I_{M},\ldots,\mu_NI_{M}\}
\end{eqnarray}
and the extended block weighting matrices
\begin{eqnarray}\label{combination_matrices}
\mathcal{C}=C\otimes I_{M}, \hspace{.5cm} \mathcal{A}=A\otimes I_{M}
\end{eqnarray}
where $\otimes$ denotes the Kronecker product operation. We further introduce the random block quantities:
\begin{eqnarray}
\bD_i&=&{\rm diag}\bigg\{\sum_{l=1}^{N}c_{l,1}\bu^*_{l,i}\bu_{l,i},\ldots,\sum_{l=1}^{N}c_{l,N}\bu^*_{l,i}\bu_{l,i}\bigg\} \label{perf_matrices}\\
\bg_i&=&\mathcal{C}^T{\rm col}\{\bu^*_{1,i}\bv_1(i),\ldots,\bu^*_{N,i}\bv_N(i)\}\label{perf_matrices2}
\end{eqnarray}
Then, we conclude from (\ref{ATC diffusion}) that the following relations hold for the error vectors:
\begin{eqnarray}
\tilde{\boldsymbol{\psi}}_{i}&=&\tilde{\bw}_{i-1}-\mathcal{M}[\bD_i\tilde{\bw}_{i-1}+\bg_i]+\gamma\mathcal{M} \partial f(\bw_{i-1})\label{psi}\\
\tilde{\bw}_{i}&=&\mathcal{A}^T\tilde{\boldsymbol{\psi}}_{i} \label{dabliu}
\end{eqnarray}
where
\begin{eqnarray}
\partial f(\bw_{i-1})\triangleq {\rm col}\{\partial f(\bw_{1,i-1}),\ldots,\partial f(\bw_{N,i-1})\}
\end{eqnarray}
We can combine (\ref{psi}) and (\ref{dabliu}) into a single recursion:
\begin{eqnarray}\label{compact_Diffusion}
\boxed{\tilde{\bw}_{i}=\mathcal{A}^T[I-\mathcal{M}\bD_i]\tilde{\bw}_{i-1}-\mathcal{A}^T\mathcal{M}\bg_i+\gamma\mathcal{A}^T\mathcal{M}\partial f(\bw_{i-1})}
\end{eqnarray}
This relation tells us how the network weight-error vector evolves over time. The relation will be the launching point for our mean-square analysis. To proceed, we introduce the following independence assumption on the regression data.

\noindent {\textbf{Assumption 1 (Independent regressors)}} {\it The regressors $\bu_{k,i}$ are temporally white and spatially independent with $R_{u,k}=\mathbb{E}\bu_{k,i}\bu_{k,i}^*>0$.}{\qedsymbol}

It follows from Assumption 1 that $\bu_{k,i}$ is independent of $\{\bw_{l,j}\}$ for all $l$ and $j\leq i-1$. Although not true in general, this assumption is common in the adaptive filtering literature since it helps simplify the analysis. Several studies in the literature, especially on stochastic approximation theory \cite{Kushner-Yin}--\cite{Sayed}, indicate that the performance expressions obtained using this assumption match well the actual performance of stand-alone filters for sufficiently small step-sizes. Therefore, we shall also rely on the following condition.

\noindent {\textbf{Assumption 2 (Small step-sizes)}} {\it The step-sizes $\{\mu_k\}$ are sufficiently small so that terms that depend on higher-order powers of $\mu_k$ can be ignored.}{\qedsymbol}

\subsection{Convergence in the Mean}

Let
\begin{eqnarray}\label{perf_matrices3}
\mathcal{D}\triangleq\mathbb{E}\bD_i={\rm diag}\bigg\{\sum_{l=1}^{N}c_{l,1}R_{u,l},\ldots,\sum_{l=1}^{N}c_{l,N}R_{u,l}\bigg\}
\end{eqnarray}
Then, taking expectations of both sides of (\ref{compact_Diffusion}) and calling upon Assumption 1, we conclude that the mean-error vector evolves according to the following dynamics:
\begin{eqnarray}\label{compact_Diffusion_Expected}
\boxed{\mathbb{E}\tilde{\bw}_{i}=\mathcal{A}^T[I-\mathcal{M}\mathcal{D}]\mathbb{E}\tilde{\bw}_{i-1}+\gamma\mathcal{A}^T\mathcal{M}\mathbb{E}\partial f(\bw_{i-1})}
\end{eqnarray}
The following theorem guarantees the asymptotic mean stability of the diffusion strategy (\ref{ATC diffusion}), and provides a closed form expression for the weight bias due to the use of the regularization term.

\noindent {\textbf{Theorem 1 (Stability in the mean)}} \textit{ Assume data model (\ref{scalar_obs}) and Assumption 1 hold. Then, for any initial condition and any choice of the matrices $A$ and $C$ satisfying (\ref{combination_coefficients2}), the diffusion strategy (\ref{ATC diffusion}) asymptotically converges in the mean if the step-sizes are chosen to satisfy:
\begin{eqnarray}\label{step_sizes}
0<\mu_k<\frac{2}{\lambda_{\max}\left({\sum_{l=1}^{N}c_{l,k}R_{u,l}}\right)} \quad\quad k=1,\ldots,N
\end{eqnarray}
where $\lambda_{\max}(X)$ denotes the maximum eigenvalue of a Hermitian positive semi-definite matrix $X$. Furthermore, as $i\rightarrow\infty$, the estimators across all nodes have biases that are given by the corresponding entries in the following bias vector:
\begin{eqnarray}\label{bias}
\hbox{bias}\triangleq\displaystyle\lim_{i\rightarrow\infty}\mathbb{E}\tilde{\bw}_i= \gamma\cdot{\cal B}\cdot \displaystyle \lim_{i\rightarrow\infty}\mathbb{E} \partial f(\bw_{i-1})
\end{eqnarray}
where 
\begin{eqnarray}
{\cal B}=\left[I-{\cal A}^T \left(I-{\cal M}\mathcal{D}\right)\right]^{-1}{\cal A}^T{\cal M}.
\end{eqnarray}
Moreover, it holds that
\begin{eqnarray}\label{bias_bound}
\|\hbox{bias}\|_{b,\infty}\leq \frac{\gamma \cdot \mu_{\max} \cdot\partial f_{\max}}{1-\delta}
\end{eqnarray}
where $\|\cdot\|_{b,\infty}$ is the block maximum norm of a vector (defined in Appendix A), $\displaystyle \mu_{\rm max}=\max_{k=1,\ldots,N} \mu_k$, $\partial f_{\rm max}=\max_i \|\partial f(\bw_{i-1})\|_{b,\infty}$ and $\delta=\rho(I-{\cal M}\mathcal{D})<1$, with $\rho(X)$ denoting the spectral radius of a matrix $X$.}

\begin{proof}
See Appendix A.
\end{proof}

\subsection{Convergence in Mean-Square}

We now examine the behavior of the steady-state mean-square deviation, $\mathbb{E}\|\tilde{\bw}_{k,i}\|^2$ as $i\rightarrow\infty$, for any of the nodes and derive conditions under which the sparse diffusion filter outperforms its unregularized version in terms of steady-state performance. In particular, we will show that, if the vector parameter $w^o$ is sufficiently sparse, then it is possible to tune the sparsity parameter $\gamma$ to achieve better performance than the standard diffusion algorithm.
Following the energy conservation framework of \cite{Lopes_Sayed,Cattivelli_Sayed} and under Assumption 1, we can establish the following variance relation:
\begin{eqnarray}\label{weighted_norm_expanded}
\mathbb{E}\|\tilde{\bw}_{i}\|^2_{\Sigma}&=&\mathbb{E}\|\tilde{\bw}_{i-1}\|^2_{\Sigma'}+ \mathbb{E}[\bg_i^*\mathcal{M}\mathcal{A}\Sigma \mathcal{A}^T\mathcal{M}\bg_i]+2\gamma\mathbb{E}\partial f(\bw_{i-1})^T\mathcal{M}\mathcal{A}\Sigma\mathcal{A}^T\left(I-{\cal M}\mathcal{D}\right)\tilde{\bw}_{i-1}\nonumber\\
&+&\gamma^2\mathbb{E}\|\partial f(\bw_{i-1})\|^2_{\mathcal{M}\mathcal{A}\Sigma\mathcal{A}^T\mathcal{M}}
\end{eqnarray}
where $\Sigma$ is any Hermitian nonnegative-definite matrix that we are free to choose, and
\begin{eqnarray}\label{Sigma'}
\Sigma'=\mathbb{E}(I-\bD_i\mathcal{M})\mathcal{A}\Sigma \mathcal{A}^T(I-\mathcal{M}\bD_i) \label{Sigma_primo}
\end{eqnarray}
Relations (\ref{weighted_norm_expanded})-(\ref{Sigma'}) can be derived directly from (\ref{compact_Diffusion}) if we compute the weighted norm of both sides of the equality  and use the fact that $\bg_i$ is independent of $\tilde{\bw}_{i-1}$ and $\bw_{i-1}$. We can rewrite (\ref{weighted_norm_expanded}) more compactly if we collect the terms depending on $\gamma$ as
\begin{eqnarray}\label{weighted_norm}
\mathbb{E}\|\tilde{\bw}_{i}\|^2_{\Sigma}=\mathbb{E}\|\tilde{\bw}_{i-1}\|^2_{\Sigma'}+ \mathbb{E}[\bg_i^*\mathcal{M}\mathcal{A}\Sigma \mathcal{A}^T\mathcal{M}\bg_i]+\phi_{\Sigma,i}(\gamma)
\end{eqnarray}
with
\begin{eqnarray}
\phi_{\Sigma,i}(\gamma)&\triangleq&\gamma\beta_{\Sigma,i}\left(\gamma-\frac{\alpha_{\Sigma,i}}{\beta_{\Sigma,i}}\right)\label{phiSigma}\\
\beta_{\Sigma,i}&=&\mathbb{E}\|\partial f(\bw_{i-1})\|^2_{\mathcal{M}\mathcal{A}\Sigma\mathcal{A}^T\mathcal{M}}\geq0 \label{beta} \\
\alpha_{\Sigma,i}&=&-2\mathbb{E}\partial f(\bw_{i-1})^T\mathcal{M}\mathcal{A}\Sigma\mathcal{A}^T\left(I-{\cal M}\mathcal{D}\right)\tilde{\bw}_{i-1}\label{alpha}
\end{eqnarray}
Moreover, setting
\begin{eqnarray}
{\cal G}=\mathbb{E}[\bg_i\bg^*_i]={\cal C}^T\cdot{\rm diag}\{\sigma_{v,1}^2R_{u,1},\ldots,\sigma_{v,N}^2R_{u,N}\}\cdot{\cal C}
\end{eqnarray}
we can rewrite (\ref{weighted_norm}) in the form
\begin{eqnarray}\label{weighted_norm2}
\mathbb{E}\|\tilde{\bw}_{i}\|^2_{\Sigma}=\mathbb{E}\|\tilde{\bw}_{i-1}\|^2_{\Sigma'}+{\rm Tr}[\Sigma \mathcal{A}^T\mathcal{M}{\cal G}\mathcal{M}\mathcal{A}]+\phi_{\Sigma,i}(\gamma)
\end{eqnarray}
where ${\rm Tr}(\cdot)$ denotes the trace operator. Let $\sigma={\rm vec}(\Sigma)$ and $\sigma'=\mbox{\rm vec}(\Sigma')$,
where the ${\rm vec}(\cdot)$ notation stacks the columns of $\Sigma$ on top of each other and ${\rm vec}^{-1}(\cdot)$ is the inverse operation. We will use interchangeably the notation $\|\tilde{w}\|^2_{\sigma}$ and $\|\tilde{w}\|^2_{\Sigma}$ to denote the same quantity $\tilde{w}^*\Sigma\tilde{w}$. Using the Kronecker product property ${\rm vec}(U\Sigma V)=(V^T\otimes U){\rm vec}(\Sigma)$, we can vectorize both sides of (\ref{Sigma'}) and conclude that (\ref{Sigma'}) can be replaced by the simpler linear vector relation: $\sigma'={\rm vec}(\Sigma')={\cal F}\sigma$,
where ${\cal F}$ is the following $N^2M^2\times N^2M^2$ matrix with block entries of size $M^2\times M^2$ each:
\begin{eqnarray}\label{matrix_F}
{\cal F}=(I\otimes I)\{I-I\otimes(\mathcal{D}\mathcal{M})-(\mathcal{D}^T\mathcal{M})\otimes I +\mathbb{E}(\bD_i^T\mathcal{M})\otimes(\bD_i\mathcal{M})\}(\mathcal{A}\otimes \mathcal{A})
\end{eqnarray}
Using the property ${\rm Tr}(\Sigma X)={\rm vec}(X^T)^T\sigma$
we can then rewrite (\ref{weighted_norm2}) as follows:
\begin{eqnarray}\label{weighted_norm3}
\mathbb{E}\|\tilde{\bw}_{i}\|^2_{\sigma}=\mathbb{E}\|\tilde{\bw}_{i-1}\|^2_{{\cal F}\sigma}+[{\rm vec}(\mathcal{A}^T\mathcal{M}{\cal G}^T\mathcal{M}\mathcal{A})]^T\sigma+\phi_{\Sigma,i}(\gamma)
\end{eqnarray}
The following theorem guarantees the asymptotic mean-square stability (i.e., convergence in the mean and mean-square sense) of the diffusion strategy (\ref{ATC diffusion}).

\noindent {\textbf{Theorem 2 (Mean-Square Stability)}} \textit{Assume the data model (\ref{scalar_obs}) and Assumption 1 hold. Then, the sparse diffusion LMS algorithm (\ref{ATC diffusion}) will be mean-square stable if the step-sizes are  sufficiently small and satisfy (\ref{step_sizes}), and the matrix ${\cal F}$ in (\ref{matrix_F}) is stable.}
\begin{proof}
See Appendix B.
\end{proof}

\noindent {\textbf{Remark 1}:} Note that the step sizes influence (\ref{matrix_F}) through the matrix $\mathcal{M}$.
Since the step-sizes are sufficiently small, we can ignore terms that depend on higher-order powers of the step-sizes. Then, we can approximate (\ref{matrix_F}) as
\begin{eqnarray}\label{approx_matrix_F}
{\cal F} \approx
(I\otimes I)\left\{
I-I\otimes ({\cal D M})-({\cal D}^T{\cal M})\otimes I+{\cal D}^T{\cal M}\otimes {\cal DM}
\right\}({\cal A}\otimes {\cal A})={\cal B}^T\otimes {\cal B}^*
\end{eqnarray}
where ${\cal B}={\cal A}^T(I-{\cal M}{\cal D})$. Now, since ${\cal A}$ is left-stochastic, it can be verified that the above ${\cal F}$ is stable if $(I-{\cal D}{\cal M})$ is stable \cite{Chen-Sayed, Sayed2}; this latter condition is guaranteed by (\ref{step_sizes}).
In summary, sufficiently small step-sizes ensure the stability of the diffusion strategy in the mean and mean-square senses. {\qedsymbol}

Taking the limit as $i \rightarrow \infty$ (assuming the step-sizes are small enough to ensure convergence to a steady-state), we deduce from (\ref{weighted_norm3}) that:
\begin{eqnarray}\label{variance_relation2}
\boxed{\displaystyle \lim_{i\rightarrow\infty}\mathbb{E}\|\tilde{\bw}_i\|^2_{(I-{\cal F})\sigma}=[{\rm vec}(\mathcal{A}^T\mathcal{M}{\cal G}^T\mathcal{M}\mathcal{A})]^T\sigma
+\gamma\beta_{\Sigma,\infty}\left(\gamma-\frac{\alpha_{\Sigma,\infty}}{\beta_{\Sigma,\infty}}\right)}
\end{eqnarray}
where
\begin{eqnarray}
\alpha_{\Sigma,\infty}&=&\lim_{i\rightarrow\infty}-2\mathbb{E}\partial f(\bw_{i-1})^T\mathcal{M}\mathcal{A}\Sigma\mathcal{A}^T\left(I-{\cal M}\mathcal{D}\right)\tilde{\bw}_{i-1}\label{limits1}\\
\beta_{\Sigma,\infty}&=&\lim_{i\rightarrow\infty}\mathbb{E}\|\partial f(\bw_{i-1})\|^2_{\mathcal{M}\mathcal{A}\Sigma\mathcal{A}^T\mathcal{M}}\label{limits2}
\end{eqnarray}
The limits in (\ref{limits1})-(\ref{limits2}) exist. Indeed, first, in Appendix C we show that $\alpha_{\Sigma,i}$ converges to $\alpha_{\Sigma,\infty}$.  Second, we also show in Appendix B that the LHS of (\ref{variance_relation2}) converges. Therefore, the term $\beta_{\Sigma,\infty}$ also exists.

Expression (\ref{variance_relation2}) is a useful result: it allows us to derive several performance metrics through the proper selection of the free weighting parameter $\sigma$ (or $\Sigma$),
as was done in \cite{Cattivelli_Sayed}. For example, the MSD for any node $k$ is defined as the steady-state value $\mathbb{E}\|\tilde{\bw}_{k,i}\|^2$, as $i\rightarrow\infty$, and can be obtained by computing $\lim_{i\rightarrow\infty}\mathbb{E}\|\tilde{\bw}_i\|^2_{T_k}$ with a block weighting matrix $T_k$ that has the $M\times M$ identity matrix at block $(k,k)$ and zeros elsewhere. Then, denoting the vectorized version of the matrix $T_k$ by $t_k={\rm vec}({\rm diag}(e_k)\otimes I_M)$, where $e_k$ is the vector whose $k$-th entry is one and zeros elsewhere, and if we select $\sigma$ in (\ref{variance_relation2}) as $\sigma_k=(I-{\cal F})^{-1}t_k$, we arrive at the MSD for node $k$:
\begin{eqnarray}\label{MSD_k}
\boxed{{\rm MSD}_k=[{\rm vec}(\mathcal{A}^T\mathcal{M}{\cal G}^T\mathcal{M}\mathcal{A})]^T(I-{\cal F})^{-1}t_k +\gamma\beta_{\Sigma_k,\infty}\left(\gamma-\frac{\alpha_{\Sigma_k,\infty}}{\beta_{\Sigma_k,\infty}}\right)}
\end{eqnarray}
The average network ${\rm MSD}_{\rm net}$ is given by:
\begin{eqnarray}\label{MSD}
{\rm MSD}_{\rm net}=\displaystyle \lim_{i\rightarrow\infty}\frac{1}{N}\sum_{k=1}^N\mathbb{E}\|\tilde{\bw}_{k,i}\|^2
\end{eqnarray}
Then, to obtain the network MSD from (\ref{variance_relation2}), the weighting matrix of $\lim_{i\rightarrow\infty}\mathbb{E}\|\tilde{\bw}_i\|^2_{T}$ should be chosen as $T=I_{MN}/N$. Let $q$ denote the vectorized version of $I_{MN}$, i.e., $q={\rm vec}(I_{MN})$, and selecting $\sigma$ in (\ref{variance_relation2}) as $\sigma=(I-{\cal F})^{-1}q/N$, the network MSD is given by:
\begin{eqnarray}\label{MSD_net}
\boxed{{\rm MSD}_{net}=\frac{1}{N}[{\rm vec}(\mathcal{A}^T\mathcal{M}{\cal G}^T\mathcal{M}\mathcal{A})]^T(I-{\cal F})^{-1}q +\frac{1}{N}\gamma\beta_{\Sigma,\infty}\left(\gamma-\frac{\alpha_{\Sigma,\infty}}{\beta_{\Sigma,\infty}}\right)}
\end{eqnarray}

\subsection{Comparison with Unregularized ATC Diffusion}

We now examine under what conditions the sparse diffusion filter (\ref{ATC diffusion}) dominates in terms of mean-square performance its unregularized counterpart when $\gamma=0$.
Considering the MSD expression (\ref{MSD_k}) at the $k$-th node, we notice that the first term on the RHS coincides with the MSD of the standard diffusion algorithm when $\gamma=0$ (compare with (48) in \cite{Cattivelli_Sayed}), whereas the second term in (\ref{MSD_k}) is due to the regularization. Then, if
\begin{eqnarray}\label{dominance_cond}
\alpha_{\Sigma_k,\infty}>0 \quad  \hbox{and} \quad 0<\gamma<\frac{\alpha_{\Sigma_k,\infty}}{\beta_{\Sigma_k,\infty}}
\end{eqnarray}
the second term on the RHS of (\ref{MSD_k}) is negative and sparse diffusion would outperform standard diffusion. The condition $\alpha_{\Sigma_k,\infty}>0$, where $\alpha_{\Sigma,i}$ is given by (\ref{alpha}), is a {\em necessary} condition to have dominance of sparse diffusion over standard diffusion. Let us examine an interpretation for the condition $\alpha_{\Sigma_k,\infty}>0$ in terms of the sparsity of the vector $w^o$. Since $f(\cdot)$ is a real-valued convex function, by the definition of subgradient it holds that
\begin{eqnarray}\label{subgrad_ineq}
f(x+y)-f(x)\geq \partial f(x)^Ty \hspace{.3cm}\Rightarrow \hspace{.3cm} -\partial f(x)^Ty\geq f(x)-f(x+y)
\end{eqnarray}
Then, choosing $x=\bw_{i-1}$ and $y=B_{\Sigma_k}(\mathds{1}\otimes w^o-\bw_{i-1})$, where
$B_{\Sigma_k}=2\mathcal{M}\mathcal{A}\Sigma_k\mathcal{A}^T\left(I-{\cal M}\mathcal{D}\right)$, we get
\begin{eqnarray}\label{Asigma}
\alpha_{\Sigma_k,\infty}&=&-2\lim_{i\rightarrow\infty}\mathbb{E}\partial f(\bw_{i-1})^T\mathcal{M}\mathcal{A}\Sigma_k\mathcal{A}^T\left(I-{\cal M}\mathcal{D}\right)\tilde{\bw}_{i-1} \nonumber\\
&\geq&\displaystyle \lim_{i\rightarrow\infty}\mathbb{E}[f(\bw_{i-1})-f(\bw_{i-1}+B_{\Sigma_k}(\mathds{1}\otimes w^o-\bw_{i-1}))]
\end{eqnarray}
If the step-sizes are sufficiently small, we can approximate $B_{\Sigma_k}\backsimeq 2\mathcal{M}\mathcal{A}\Sigma_k\mathcal{A}^T$, neglecting the second term that depends on $\{\mu_k^2\}$. Then, we have
\begin{eqnarray}\label{bar_w_i}
\bar{\bw}_{i}\triangleq\bw_{i-1}+B_{\Sigma_k}(\mathds{1}\otimes w^o-\bw_{i-1})\backsimeq\bw_{i-1}-2\mathcal{M}\mathcal{A}\Sigma_k\mathcal{A}^T(\bw_{i-1}-\mathds{1}\otimes w^o)
\end{eqnarray}
At convergence, the vector $\bw_{i-1}$ fluctuates close to $\mathds{1}\otimes w^o$. Now, since $\Sigma_k\geq0$, expression (\ref{bar_w_i}) can be interpreted as a gradient descent update minimizing the function $\|\bw-\mathds{1}\otimes w^o\|^2_{\mathcal{A}\Sigma_k\mathcal{A}^T}$, yielding for small step-sizes a vector $\bar{\bw}_{i}$ that is closer to $\mathds{1}\otimes w^o$ than $\bw_{i-1}$. If $\mathds{1}\otimes w^o$ is sparse, the non-zero elements (NZ set) of the vector are in general much less in number than the zero elements (Z set). Then, the gradient update in (\ref{bar_w_i}) helps move the components of the vector $\bar{\bw}_{i}$ that belong to the Z set closer to zero. Intuitively, if the Z set is larger than the NZ set, $\bar{\bw}_{i}$ will be more sparse than $\bw_{i-1}$. Thus, considering (\ref{Asigma}) at convergence, since the function $f(w)$ measures the sparsity of the vector $w$, it is expected that
\begin{eqnarray}
\displaystyle\lim_{i\rightarrow\infty}\mathbb{E}[f(\bw_{i-1})-f(\bar{\bw}_i)] >0
\end{eqnarray}
since $\bar{\bw}_i$ is likely to be more sparse than $\bw_{i-1}$. Consequently, the condition $\alpha_{\Sigma_{k},\infty}>0$ is likely to be true. Therefore, by properly selecting the sparsity coefficient $\gamma$ to satisfy (\ref{dominance_cond}), the sparse diffusion algorithm will yield better MSD than the standard diffusion algorithm at each node. On the other hand, if $w^o$ is not sparse, condition (\ref{Asigma}) in general would not be true and the sparse diffusion algorithm will perform worse than standard diffusion.

\subsection{Adaptation of the Regularization Parameter}

To endow networks with the capability to adaptively exploit and track the sparsity of the system model, we now propose a systematic approach to choosing the regularization parameter $\gamma$ in an adaptive fashion. We thus allow the sparsity parameter to be iteration dependent, i.e., $\gamma=\gamma_i$. Following similar steps as in Section III.B, we can replace
(\ref{weighted_norm2}) with the conditional relation:
\begin{eqnarray}\label{conditioned_weighted_norm}
\mathbb{E}\left[\|\tilde{\bw}_{i}\|^2_{\Sigma}|w_{i-1}\right]=\|\tilde{w}_{i-1}\|^2_{\Sigma'}+{\rm Tr}[\Sigma \mathcal{A}^T\mathcal{M}{\cal G}^T\mathcal{M}\mathcal{A}]+
\phi_{\Sigma,i}(\gamma_i)
\end{eqnarray}
where $\Sigma'$ is given by (\ref{Sigma'}) and
\begin{eqnarray}
\phi_{\Sigma,i}(\gamma_i)&=&\gamma_i\beta_{\Sigma,i}\left(\gamma_i-\frac{\alpha_{\Sigma,i}}{\beta_{\Sigma,i}}\right) \label{instant_phi}\\
\beta_{\Sigma,i}&=&\|\partial f(w_{i-1})\|^2_{\mathcal{M}\mathcal{A}\Sigma\mathcal{A}^T\mathcal{M}}\geq0\label{beta_cond} \\
\alpha_{\Sigma,i}&=&-2\partial f(w_{i-1})^T\mathcal{M}\mathcal{A}\Sigma\mathcal{A}^T\left[I-{\cal M}\mathcal{D}\right]\tilde{w}_{i-1}\label{alpha_cond}
\end{eqnarray}
Thus, letting $\Sigma=I$ and if $\phi_{\Sigma,i}(\gamma_i)<0$, the sparse diffusion algorithm will outperform the standard diffusion algorithm in terms of the instantaneous MSD. The condition $\phi_{\Sigma,i}(\gamma_i)<0$ is satisfied when
\begin{eqnarray}\label{dominance_cond2}
\alpha_{\Sigma,i}>0 \quad  \hbox{and} \quad 0<\gamma_{i}<\frac{\alpha_{\Sigma,i}}{\beta_{\Sigma,i}}
\end{eqnarray}
Since $\phi_{\Sigma,i}(\gamma_i)$ in (\ref{instant_phi}) is quadratic in $\gamma_i$, we can choose the optimal parameter that minimizes (\ref{instant_phi}) as:
\begin{eqnarray}\label{opt_rho}
\gamma^o_{i}=\max\left\{0,\frac{\alpha_{\Sigma,i}}{2\beta_{\Sigma,i}}\right\}
\end{eqnarray}
Now, exploiting the small step-sizes assumption in (\ref{alpha_cond}), we consider the following approximation:
\begin{eqnarray}
\alpha_{\Sigma,i}\simeq-2\partial f(w_{i-1})^T\mathcal{M}\mathcal{A}\mathcal{A}^T\tilde{w}_{i-1}\label{alpha_approx}
\end{eqnarray}
An approximate expression for the sparsity parameter in (\ref{opt_rho}) is then given by:
\begin{eqnarray}\label{opt_rho_network}
\gamma^o_{i}\simeq\max\left\{0,\frac{-\partial f(w_{i-1})^T\mathcal{M}\mathcal{A}\mathcal{A}^T\tilde{w}_{i-1}}{\|\partial f(w_{i-1})\|^2_{\mathcal{M}\mathcal{A}\mathcal{A}^T\mathcal{M}}}\right\}
\end{eqnarray}

\noindent {\textbf{Remark 2}:} The rule (\ref{opt_rho_network}) cannot be directly used due to the presence of the true parameter vector $w^o$ in $\tilde{w}_{i-1}$, which is unknown to the nodes in the network. Furthermore, the update (\ref{opt_rho_network}) depends on data coming from all nodes. However, in the sequel we propose some useful approximations that allow the local computation of the regularization parameter. {\qedsymbol}

First, we notice that the regularization parameter (\ref{opt_rho_network}) depends on the combination matrix $\mathcal{A}$, which influences how the nodes perform the combination step in (\ref{ATC diffusion}). This step helps improve the quality of the node's estimate $w_{k,i}$ by reducing the effect of the measurement and gradient noises but,
it generally has a marginal effect on the sparse recovery capability of the algorithm. The regularization function appears instead inside the adaptation step in (\ref{ATC diffusion}). Thus, to simplify expression (\ref{opt_rho_network}), we consider the case in which we want to select $\gamma_i$ under the condition that ${\cal A}=I$, i.e., no cooperation is performed among the nodes. In this case, the following relations hold:
\begin{eqnarray}
\beta_{\Sigma,i}&\simeq&\sum_{k=1}^N \mu_k^2 \|\partial f(w_{k,i-1})\|^2 \label{beta_cond_approx}\\
\alpha_{\Sigma,i}&\simeq&-2\sum_{k=1}^N\mu_k \partial f(w_{k,i-1})^T\tilde{w}_{k,i-1}\label{alpha_cond_approx}
\end{eqnarray}
Let $x=w_{k,i-1}$ and $y=w^o-w_{k,i-1}$. Using (\ref{subgrad_ineq}), we find that
\begin{eqnarray}\label{alpha_cond_approx2}
\alpha_{\Sigma,i}\simeq-2\sum_{k=1}^N\mu_k \partial f(w_{k,i-1})^T\tilde{w}_{k,i-1}\geq2\sum_{k=1}^N\mu_k [f(w_{k,i-1})-f(w^o)]\label{alpha_cond_approx}
\end{eqnarray}
In practice, some prior knowledge about the sparsity of the true vector $w^o$ is often available. For example, the $\ell_1$-norm of $w^o$ can be upper bounded by some constant value \cite{Tibshirani}. In this work, we assume that
\begin{eqnarray}\label{eta}
f(w^o)\leq\eta,
\end{eqnarray}
for some given positive constant $\eta$. Using (\ref{eta}) in (\ref{alpha_cond_approx2}), we get
\begin{eqnarray}\label{alpha_cond_approx3}
\alpha_{\Sigma,i}\geq2\sum_{k=1}^N\mu_k [f(w_{k,i-1})-\eta]
\end{eqnarray}
and, using (\ref{beta_cond_approx}) and (\ref{alpha_cond_approx3}), the regularization parameter in (\ref{opt_rho}) can instead be approximated as:
\begin{eqnarray}\label{opt_rho2}
\gamma^o_{i}=\max\left\{0,\frac{\sum_{k=1}^N\mu_k [f(w_{k,i-1})-\eta]}{\sum_{k=1}^N \mu_k^2 \|\partial f(w_{k,i-1})\|^2}\right\}
\end{eqnarray}

\noindent {\textbf{Remark 3}:} The update (\ref{opt_rho2}) still depends on data coming from all nodes in the network. However, we can replace (\ref{opt_rho2}) with a local rule where each node computes its own $\rho^o_{k,i}$ from data received from its neighbors only, say,
\begin{eqnarray}\label{opt_rho4}
\displaystyle \gamma^o_{k,i}=\max\left\{0,\frac{\sum_{l\in{\cal N}_k}\mu_l [f(w_{l,i-1})-\eta]}{\sum_{l\in{\cal N}_k} \mu_l^2 \|\partial f(w_{l,i-1})\|^2}\right\}
\end{eqnarray}
In the simulation section, we will check the performance of the sparse diffusion strategy using (\ref{opt_rho4}). {\qedsymbol}

We summarize below the sparse diffusion strategy with adaptive regularization. The complexity of this strategy is $O(4M)$, which is the same complexity as standard stand-alone LMS adaptation.

\vspace{.3cm}
\begin{algorithm}
\caption*{\textbf{ATC sparse diffusion LMS with adaptive regularization}}
\vspace{.3cm}
Start with $w_{k,-1}=0$ for all $k$. Given non-negative real coefficients $\{a_{l,k},c_{l,k}\}$ satisfying (\ref{combination_coefficients2}), for each time $i\geq0$ and for each node $k$, repeat:
\begin{eqnarray}\label{ATC diffusion_adaptive_regularization}
\begin{cases}
\displaystyle \gamma^o_{k,i}=\max\left\{0,\frac{\sum_{l\in{\cal N}_k}\mu_l [f(w_{l,i-1})-\eta]}{\sum_{l\in{\cal N}_k} \mu_l^2 \|\partial f(w_{l,i-1})\|^2}\right\}  \hspace{5.2cm} \hbox{(sparsity control)}  \\
\psi_{k,i}=w_{k,i-1}+\mu_k \displaystyle \sum_{l\in \mathcal{N}_k}c_{l,k} u^*_{l,i}[d_{l}(i)-u_{l,i}w_{k,i-1}]-\mu_k\gamma^o_{k,i} \partial f(w_{k,i-1}) \hspace{1cm} \hbox{(adaptation step)} \\
w_{k,i}=\displaystyle\sum_{l \in {\cal N}_k}a_{l,k}\psi_{l,i} \hspace{9.2cm} \hbox{(diffusion step)}
\end{cases}
\end{eqnarray}
\end{algorithm}

\noindent {\textbf{Remark 4}:} Equation (\ref{opt_rho4}) indicates that, in order to ensure superiority of the sparse diffusion strategy, the construction (\ref{opt_rho4}) is triggered only if $\sum_{l\in{\cal N}_k}\mu_l [f(w_{l,i-1})-\eta]>0$, otherwise, $\gamma^o_{i}=0$. The performance of the sparse diffusion strategy depends on how close the upper bound $\eta$ is to the right value. In the simulation section, we will check the robustness of the regularized diffusion algorithm to misspecified values of $\eta$. {\qedsymbol}

\section{Simulation Results}

In this section, we provide some numerical examples to illustrate the performance of the sparse diffusion algorithm. In the first example, we compare the performance of the sparse diffusion strategy with respect to standard diffusion, considering fixed values of the regularization parameter $\gamma$. The second example shows the benefits of adapting the sparsity parameter according to (\ref{opt_rho4}).
\begin{figure}[t]
\centering
\includegraphics[scale=0.6]{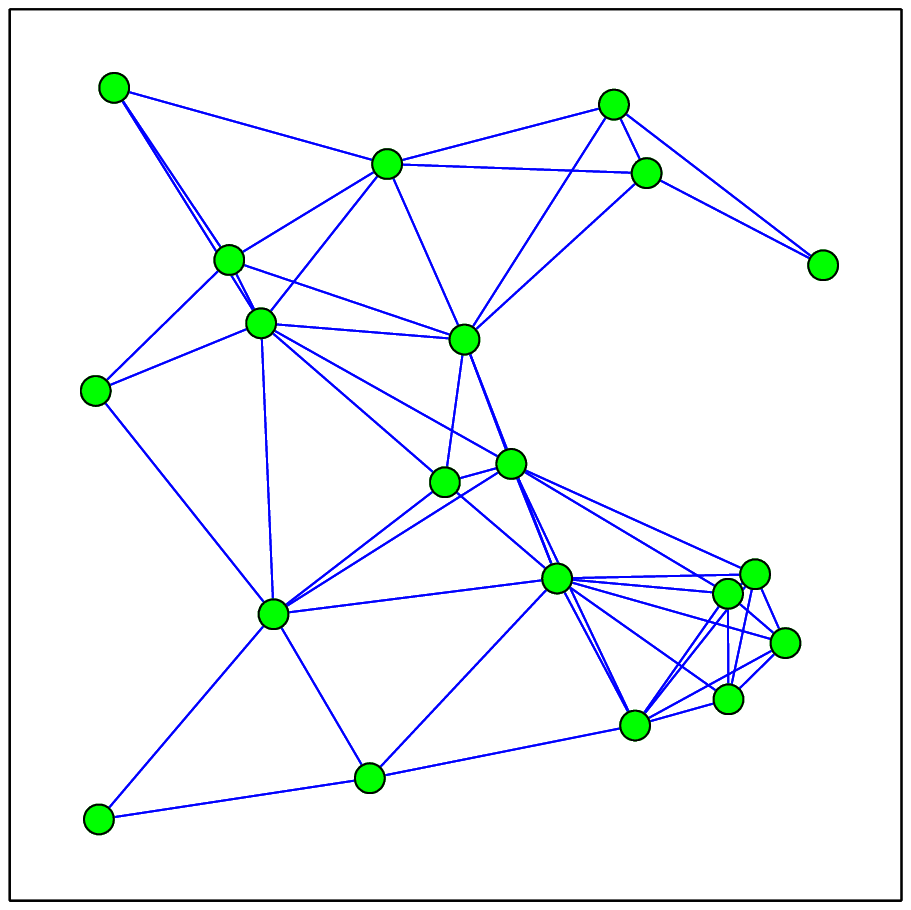}
\hspace{.35cm}\includegraphics[scale=0.5]{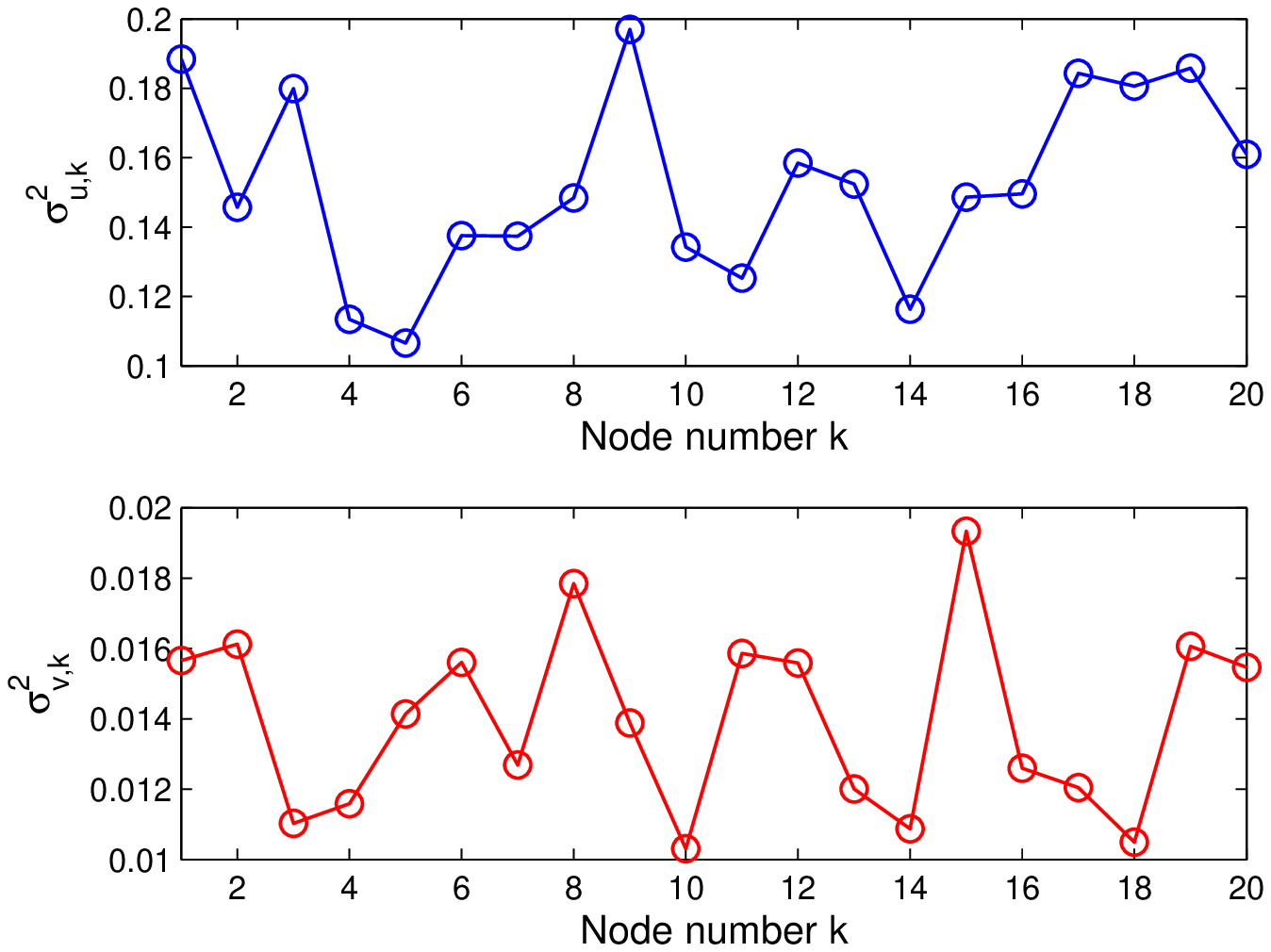}
  \caption{Network topology (left), noise variances (right, bottom) and regressor variances (right, top).}\label{net}
\end{figure}

\begin{figure}[b]
\centering
\includegraphics[scale=0.61]{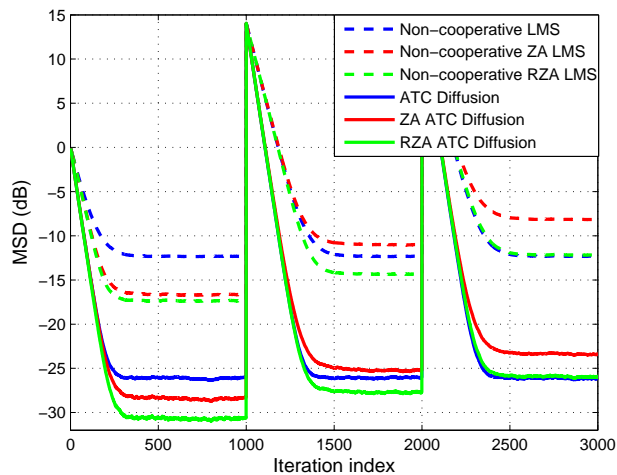}
  \caption{Transient network MSD for the non-cooperative approaches LMS, ZA-LMS \cite{Chen-Gu-Hero}, RZA-LMS \cite{Chen-Gu-Hero}, and the diffusion techniques ATC \cite{Cattivelli_Sayed}, ZA-ATC given by
  (\ref{ATC diffusion})-(\ref{sign}), RZA-ATC given by (\ref{ATC diffusion})-(\ref{update_rew_l1norm}).}\label{Comparison}
\end{figure}

{\it Numerical Example 1 : Performance :} We consider a connected network composed of 20 nodes. The topology of the network is shown in Fig. \ref{net}. The regressors $\bu_{k,i}$ have size $M=50$ and are zero-mean white Gaussian distributed with covariance matrices $R_{u,k}=\sigma_{u,k}^2I_M$, with $\sigma_{u,k}^2$ shown on the top right side of Fig. \ref{net}. The background white noise power $\sigma_{u,k}^2$ of each node is depicted on the bottom right side of Fig. \ref{net}. The first example aims to show the tracking and steady-state performance for the sparse diffusion algorithm. In Fig. \ref{Comparison}, we report the learning curves in terms of network MSD for 6 different adaptive filters: ATC diffusion LMS \cite{Cattivelli_Sayed}, ZA-ATC diffusion described by (\ref{ATC diffusion}) and (\ref{sign}) and RZA-ATC diffusion described by (\ref{ATC diffusion}) and (\ref{update_rew_l1norm}), and the non-cooperative approach from \cite{Chen-Gu-Hero}.
The simulations use a value of $\mu=0.1$ and the results are averaged over 100 independent experiments. The sparsity parameters are set to $\gamma=5\times10^{-3}$ for ZA-LMS, $\gamma=0.7\times10^{-3}$ for RZA-LMS, $\gamma_{ZA}=10^{-3}$ for ZA-ATC, $\gamma_{RZA}=0.25\times10^{-3}$ for RZA-ATC, and $\varepsilon=0.1$. In this simulation, we consider diffusion algorithms without measurement exchange, i.e., $C=I$, and a combination matrix $A$ that simply averages the estimates from the neighborhood such that $a_{l,k}=1/|{\cal N}_k|$ for all $l$. Initially, only one of the 50 elements of $w^o$ is set equal to one while the others are equal to zero, making the system very sparse. After 1000 iterations, 25 elements are randomly selected and set equal to 1, making the system have a sparsity ratio of $25/50$. After 2000 iterations, all the elements are set equal to 1, leaving a completely non-sparse system. As we see from Fig. \ref{Comparison}, when the system is very sparse both ZA-ATC and RZA-ATC yield better steady-state performance than standard diffusion. The RZA-ATC outperforms ZA-ATC thanks to reweighted regularization. When the vector $w^o$ is only half sparse, the performance of ZA-ATC deteriorates, performing worse than standard diffusion, while RZA-ATC has the best performance among the three diffusion filters. When the system is completely non-sparse, the RZA-ATC still performs comparably to the standard diffusion filter. We also notice the gain of diffusion schemes with respect to the non-cooperative approaches from \cite{Chen-Gu-Hero}.

\begin{figure}[t]
\centering
\includegraphics[scale=0.53]{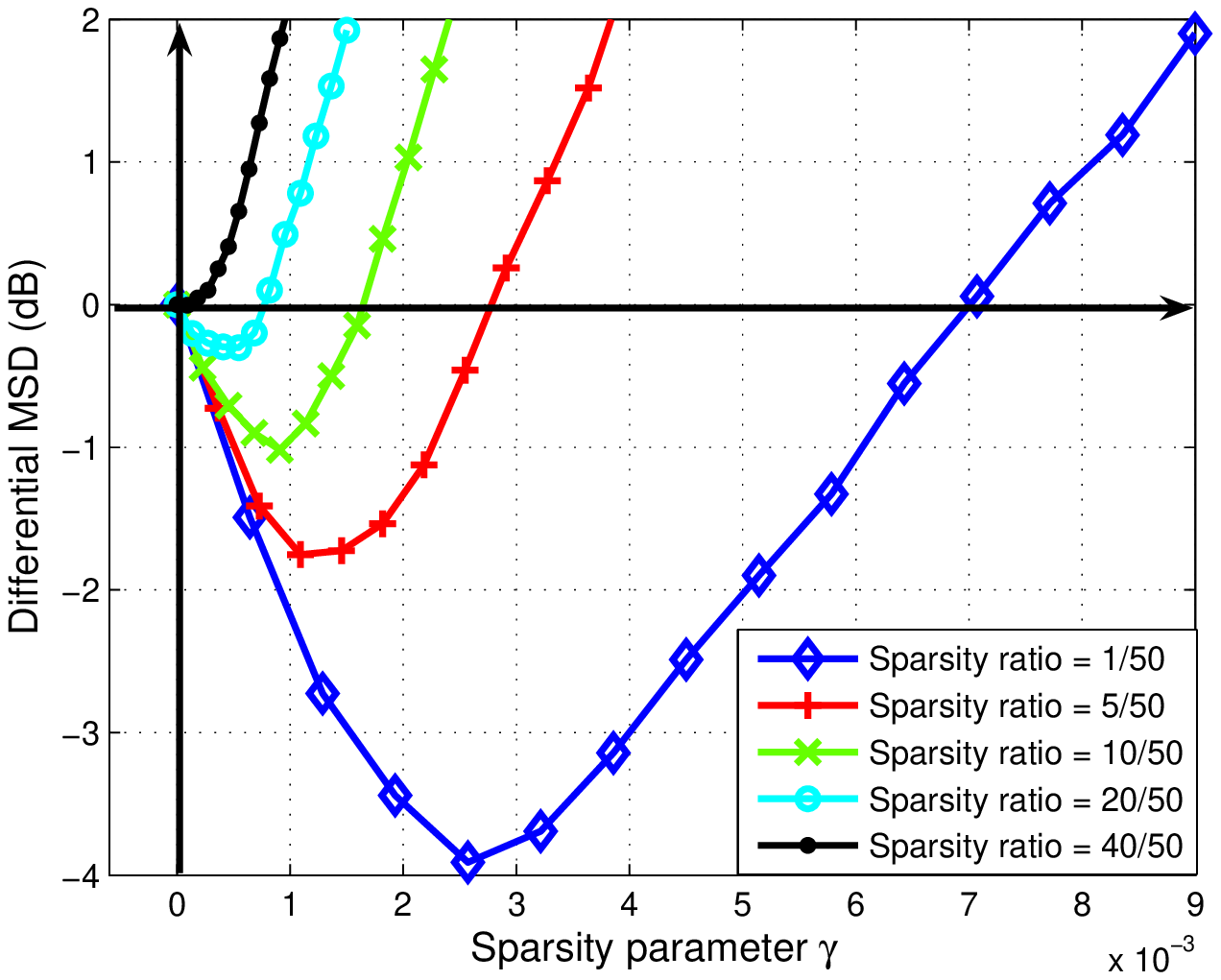}
\hspace{.45cm}\includegraphics[scale=0.53]{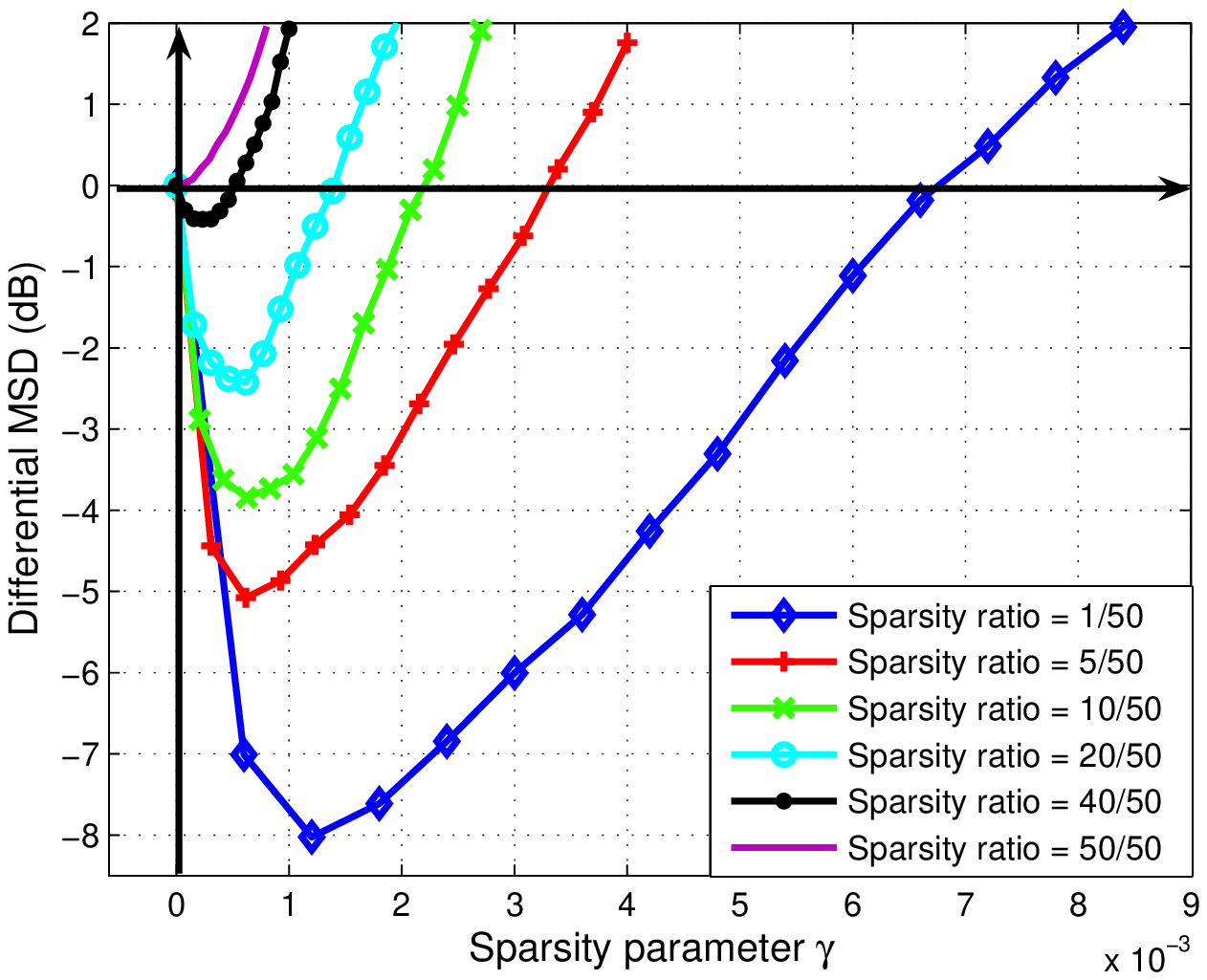}
\caption{Differential MSD versus sparsity parameter $\gamma$ for ZA-ATC Diffusion LMS (left) and for RZA-ATC Diffusion LMS (right), for different degrees of system sparsity.}\label{DMSD_ZA}
\end{figure}
The theoretical derivations in Section III showed that it is possible to select the regularization parameter $\gamma$ in order to have dominance in terms of MSD of the ATC-SD filter with respect to the unregularized diffusion algorithm.
To quantify the effect of the sparsity parameter $\gamma$ on the performance of the ATC-SD filters with respect to different degrees of system sparsity, we consider two additional examples. In Fig. \ref{DMSD_ZA} (left), we show the behavior of the difference (in dB) between the network MSD of ATC-ZA and standard diffusion versus $\gamma$, for different sparsity degrees of $w^o$. We consider the same settings of the previous simulation and the results are averaged over 100 independent experiments and over 100 samples after convergence. As we can see from Fig. \ref{DMSD_ZA} (left), reducing the sparsity of $w^o$, the interval of $\gamma$ values that yields a gain for ATC-ZA with respect to standard diffusion becomes smaller, until it reduces to zero when the system is not sparse enough.
Different update functions may affect differently the steady-state performance of the ATC-SD algorithm. Thus, in Fig. \ref{DMSD_ZA} (right), we repeat the same experiment considering the ATC-RZA algorithm. As we can see, thanks to the reweighted regularization in (\ref{update_rew_l1norm}), ATC-RZA gives better performance than ZA-ATC and yields a performance loss with respect to standard diffusion, for any $\gamma$, only when the vector $w^o$ is completely non-sparse.

Finally, we compare our proposed sparse diffusion schemes with the sparsity promoting adaptive algorithms for distributed learning recently proposed in \cite{Chouvardas-Slavakis-Kopsinis-Theodoridis} and in \cite{Liu-Li-Zhang}. At the best of our knowledge, the works in \cite{Chouvardas-Slavakis-Kopsinis-Theodoridis} and \cite{Liu-Li-Zhang} are the only two present in the literature that exploit sparsity processing data both in an adaptive and distributed fashion.
\begin{figure}[t]
\centering
\includegraphics[scale=0.59]{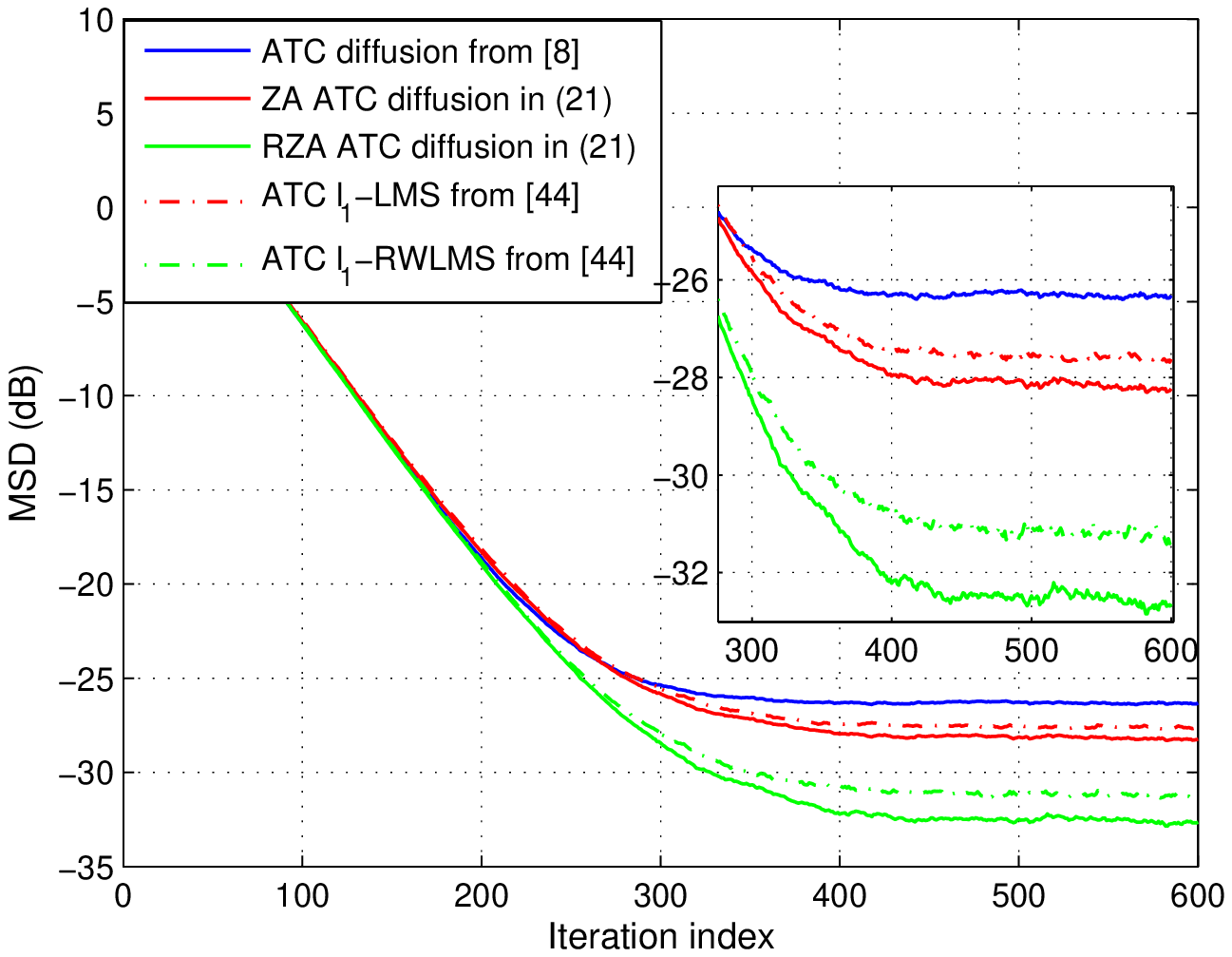}
\hspace{.45cm}\includegraphics[scale=0.59]{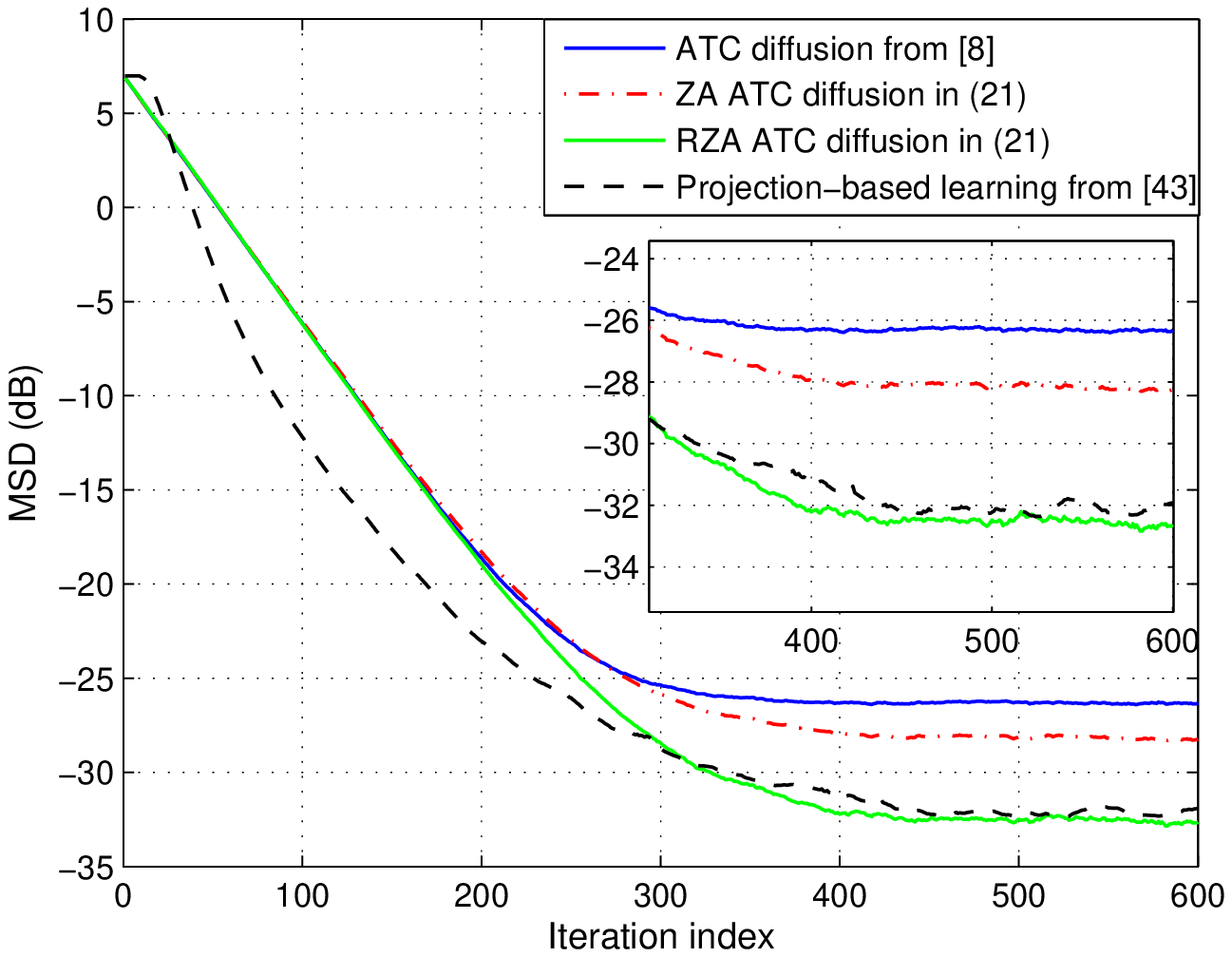}
  \caption{ (Left) Transient network MSD for the diffusion techniques ATC \cite{Cattivelli_Sayed}, ZA-ATC described by (\ref{ATC diffusion}) and (\ref{sign}), RZA-ATC described by (\ref{ATC diffusion}) and (\ref{update_rew_l1norm}), and the sparse diffusion algorithms from \cite{Liu-Li-Zhang}. (Right) Transient network MSD for the diffusion techniques ATC \cite{Cattivelli_Sayed}, ZA-ATC described by (\ref{ATC diffusion}) and (\ref{sign}), RZA-ATC described by (\ref{ATC diffusion}) and (\ref{update_rew_l1norm}), and the projection based distributed learning technique from \cite{Chouvardas-Slavakis-Kopsinis-Theodoridis}.}\label{Comparison3}
\end{figure}
In Fig. \ref{Comparison3} (left), we compare the steady-state performance, averaged over 100 independent simulations, of five adaptive filters: ATC diffusion LMS \cite{Cattivelli_Sayed}, ZA-ATC diffusion described by (\ref{ATC diffusion}) and (\ref{sign}), RZA-ATC diffusion described by (\ref{ATC diffusion}) and (\ref{update_rew_l1norm}), the ATC $\ell_1$-LMS and the ATC $\ell_1$-RWLMS algorithms from \cite{Liu-Li-Zhang}. We consider a vector parameter $w^o$ with only 5 elements set equal to one, which have been randomly chosen, leading to a sparsity ratio of 5/50. The sparsity parameters are set to $\gamma_{ZA}=10^{-3}$ for ZA-ATC, $\gamma_{RZA}=0.7\times10^{-3}$ for RZA-ATC, and $\varepsilon=0.1$, for both our methods and the algorithms from \cite{Liu-Li-Zhang}. The other settings are the same of the previous simulation, except that in this simulation the combination coefficients in (\ref{ATC diffusion}) are chosen as $c_{l,k}=1/|{\cal N}_k|$ for all $l$, thus leading to ZA-ATC and RZA-ATC diffusion algorithms with measurement exchange. As we can notice from Fig. \ref{Comparison3} (left), the proposed methods outperform the algorithms from \cite{Liu-Li-Zhang} in terms of steady-state MSD. In Fig. \ref{Comparison3} (right), we compare the transient network MSD of four adaptive filters: ATC diffusion LMS \cite{Cattivelli_Sayed}, ZA-ATC diffusion described by (\ref{ATC diffusion}) and (\ref{sign}), RZA-ATC diffusion described by (\ref{ATC diffusion}) and (\ref{update_rew_l1norm}), and the projection based sparse learning from \cite{Chouvardas-Slavakis-Kopsinis-Theodoridis}. The settings of the ZA-ATC and RZA-ATC diffusion algorithms are the same of the previous simulation, whereas the parameters of the algorithm from \cite{Chouvardas-Slavakis-Kopsinis-Theodoridis} are chosen in order to have similar steady-state MSD with respect to the RZA-ATC diffusion method. Using the same notation adopted in \cite{Chouvardas-Slavakis-Kopsinis-Theodoridis}, the parameters of the projection based filter are: $\varepsilon=1.3\times\max_k(\sigma_{v,k})$; $\mu_n=0.06\times\mathcal{M}_n$; the radius of the weighted $\ell_1$ ball is equal to $\|w^o\|_0=5$ (i.e., the correct sparsity level); $\tilde{\varepsilon}_n=0.02$; $\alpha=0.99$ for $i<300$ and $\alpha=0.8$ for $i>300$; the number of hyperslabs used per time update equals to $q=10$. From Fig. \ref{Comparison3} (right), it is possible to notice how the projected based method has a larger convergence rate with respect to the RZA ATC diffusion method. This positive feature is paid in terms of computational complexity. Indeed, while our methods have an LMS type complexity $O(3M)$, the projection-based method from \cite{Chouvardas-Slavakis-Kopsinis-Theodoridis} has a complexity equal to $O(M (3+q+\log M))$, due to the presence of $q$ projections onto the hyperslabs and 1 projection on the weighted $\ell_1$ ball per iteration.

\begin{figure}[t]
\centering
\includegraphics[scale=0.54]{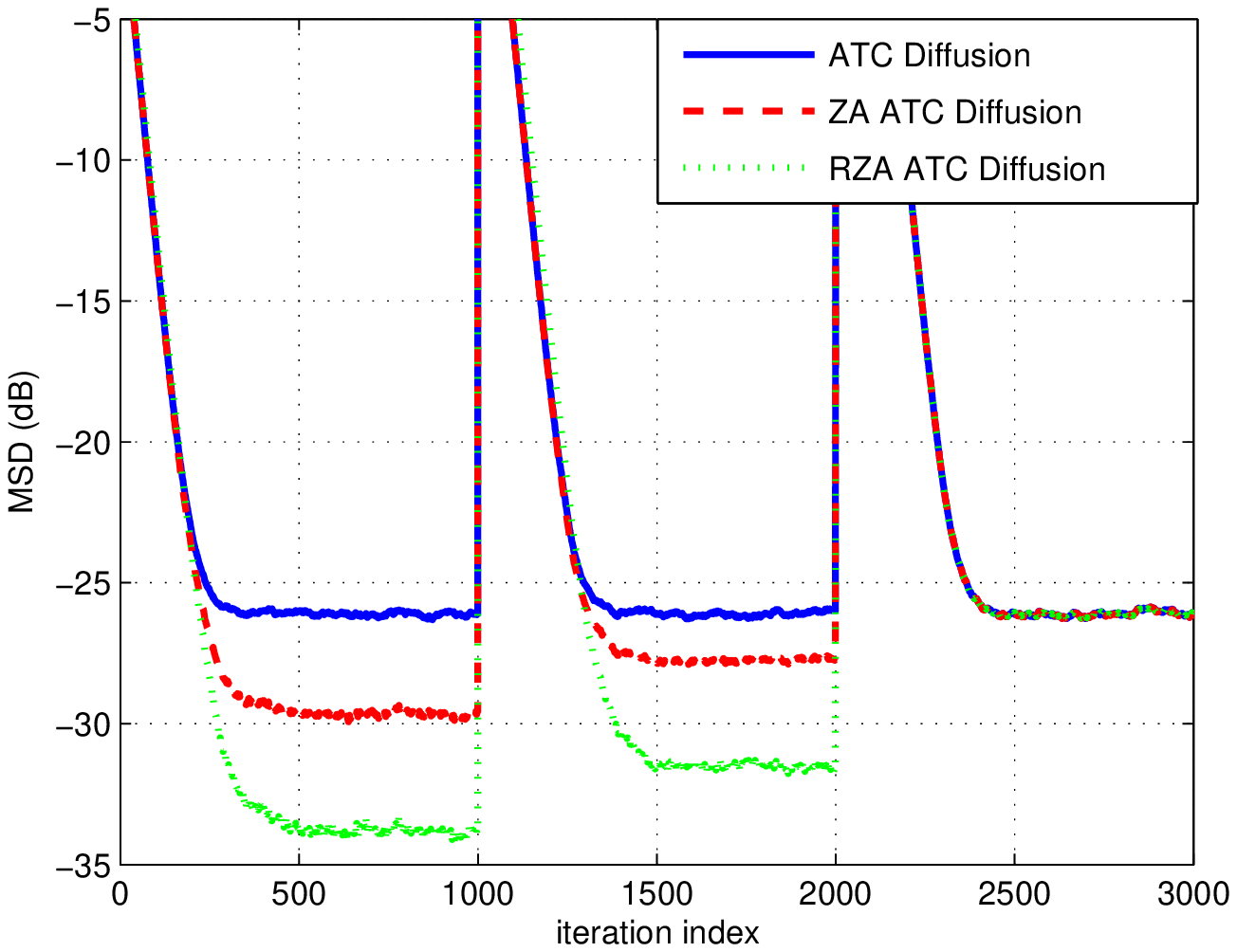}
\hspace{.35cm}\includegraphics[scale=0.54]{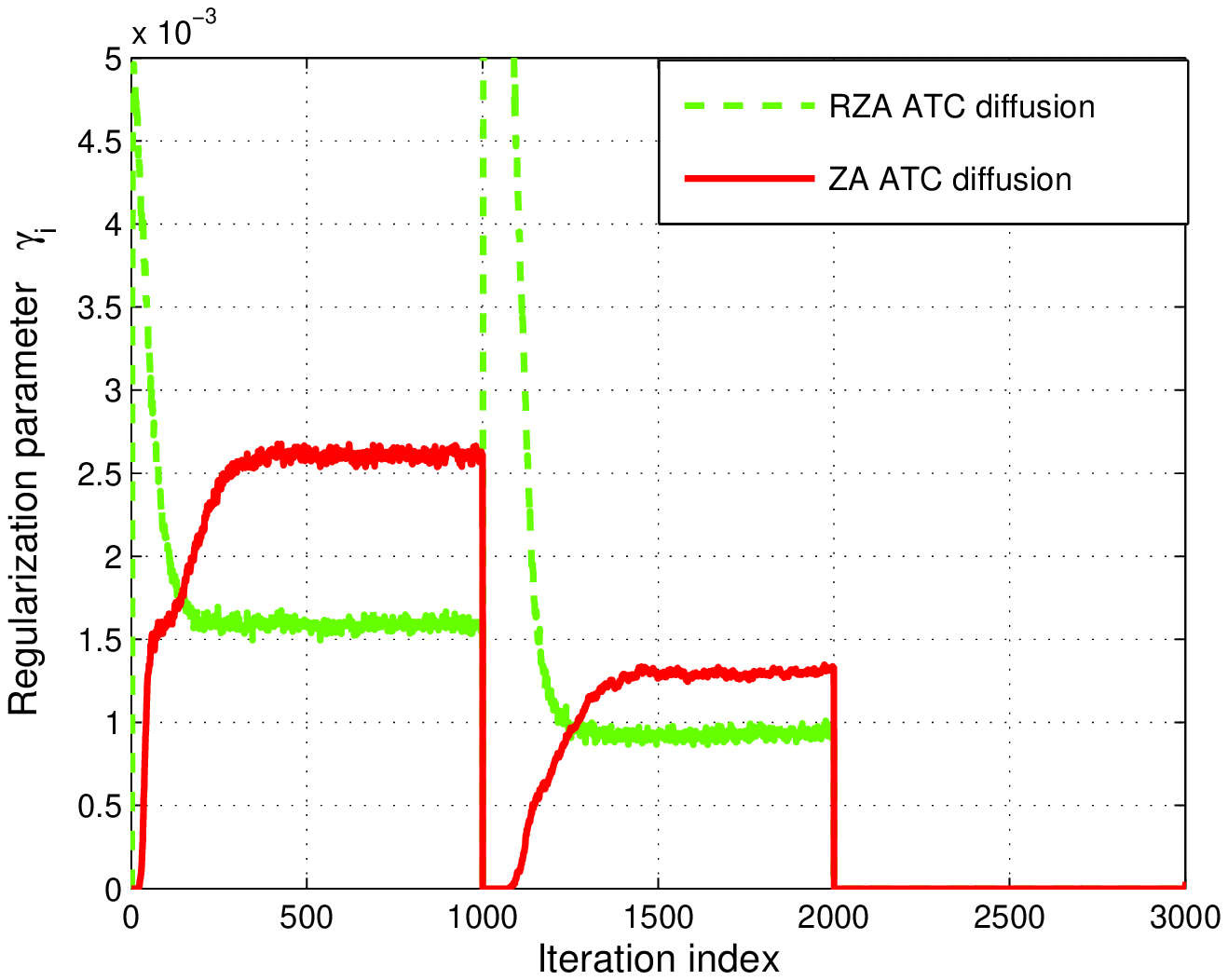}
  \caption{(Left) Transient network MSD for the the diffusion techniques ATC \cite{Cattivelli_Sayed}, ZA-ATC described by (\ref{ATC diffusion}) and (\ref{sign}), RZA-ATC described by (\ref{ATC diffusion}) and (\ref{update_rew_l1norm}) with adaptive selection of the regularization parameter $\gamma_i$. (Right) Temporal behavior of the regularization parameter $\gamma_i$ evaluated through the adaptive relation (\ref{opt_rho4}) for ZA-ATC diffusion (solid) and RZA-ATC diffusion (dashed).}\label{Comparison2}
\end{figure}

{\it Numerical Example 2 - Adaptation of the Regularization Parameter:} In this example, we consider the same network shown in Fig. \ref{net} and the same setting of the previous simulation for the regression data and additive noise. The first example aims to show the tracking and steady-state performance of the ATC-SD algorithm with adaptive regularization.
In Fig. \ref{Comparison2} (left), we report the learning curves in terms of network MSD for 3 different adaptive filters: ATC diffusion LMS \cite{Cattivelli_Sayed}, ZA-ATC diffusion described by (\ref{ATC diffusion}) and (\ref{sign})) and RZA-ATC diffusion described by (\ref{ATC diffusion}) and (\ref{update_rew_l1norm}), when the regularization parameter $\gamma_i$ is chosen locally at each node according to the adaptive rule (\ref{opt_rho4}). The simulations use a value of $\mu=0.1$ and the results are averaged over 100 independent experiments. The approximation parameter for RZA-ATC diffusion in (\ref{update_rew_l1norm}) is chosen equal to $\varepsilon=0.1$.
\begin{figure}[b]
\centering
\includegraphics[scale=0.58]{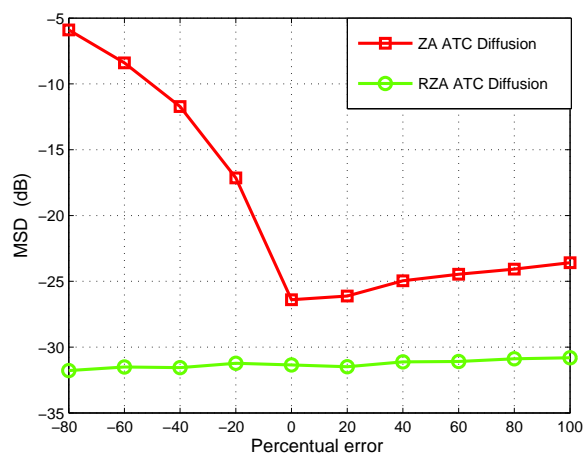}
  \caption{Sensitivity of ZA-ATC diffusion and RZA-ATC diffusion to misspecifications of the trigger parameter $\eta$.}\label{MSD_error}
\end{figure}
Initially, only one of the 50 elements of $w^o$ is set equal to one while the others are equal to zero, making the system very sparse. After 1000 iterations, 5 elements are randomly selected and set equal to 1, making the system have a sparsity ratio of $5/50$. After 2000 iterations, all the elements are set equal to 1, leaving a completely non-sparse system. The upper bound $\eta$ in (\ref{eta}), used to evaluate the sparsity parameter in (\ref{opt_rho4}), is set to $\eta=\|w^o\|_1$ and varies in time according to the different choices of $w^o$. As we can see from Fig. \ref{Comparison2} (left), when the system is very sparse both ZA-ATC and RZA-ATC yield better steady-state performance than standard diffusion. The RZA-ATC outperforms ZA-ATC thanks to the reweighted regularization. When the vector $w^o$ is less sparse, the performance of ZA-ATC deteriorates, getting closer to standard diffusion, while RZA-ATC still guarantees a large gain. When the system is completely non-sparse, the three filters have the same performance. To see the effect of different sparsity ratios of the vector $w^o$ on the choice of the regularization, in Fig. \ref{Comparison2} (right) we show the average behavior of the parameter $\gamma^o_i$ evaluated according to (\ref{opt_rho4}), for ZA-ATC diffusion and RZA-ATC diffusion, averaged across nodes over 100 independent realizations. As we can see, the system reacts to different sparsity ratios of the vector $w^o$, adjusting accordingly the regularization parameter $\gamma^o_i$ in order to improve the performance of the ATC-SD strategy with respect to the unregularized algorithm. From Fig. \ref{Comparison2} (right), it is interesting to note how the regularization parameter converges close to the minimum of the Differential MSD plotted in Fig. \ref{DMSD_ZA} for both ZA-ATC and RZA-ATC. In particular, $\gamma^o_i$ is forced to zero when the vector $w^o$ is totally non-sparse, leading to the same performance of the standard diffusion algorithm.

Since the adaptive update of the sparsity parameter $\gamma_i^o$ in (\ref{opt_rho4}) depends on the selection of the trigger $\eta$, which depends on some available prior knowledge on the sparsity level of $w^o$, it is important to check the sensitivity of the ATC-SD algorithm to misspecified values of $\eta$.
Thus, in Fig. \ref{MSD_error}, we report the average behavior of the MSD, for ZA-ATC diffusion and RZA-ATC diffusion, versus a percentual error on the specification of the true trigger value $\eta$. The settings are the same of the previous simulation and the results are averaged over 100 independent experiments and over 100 samples after convergence. We consider a vector parameter $w^o$ with only 5 elements set equal to one, which have been randomly chosen, leading to a sparsity ratio of 5/50. In this case, the true value for the trigger parameter $\eta$ would be equal to $\|w^o\|_1=5$. The regularization parameter $\gamma_i$ is chosen locally at each node according to the adaptive rule (\ref{opt_rho4}). As we can notice from Fig. \ref{MSD_error}, the ZA-ATC diffusion algorithm is very sensitive to misspecified values of $\eta$, especially in the case of under-estimation of the trigger parameter. Indeed, by under-estimating the value of $\eta$, the system would try to increase the sparsity parameter $\gamma_i$, in order to make the solution more sparse. Thus, as we notice from Fig. \ref{MSD_error}, being the true vector $w^o$ not sparse enough with respect to the selection of the trigger $\eta$, the system determines an increment of the bias that strongly affects the performance. On the contrary, from Fig. \ref{MSD_error}, we notice how the RZA-ATC diffusion algorithm is robust to errors in the selection of the trigger parameter $\eta$. This benefit is again due to regularization, whose presence reduces the magnitude of the bias, improving the estimation capabilities of the algorithm and relaxing the choice of the system parameters.

\section{Conclusion}

In this paper we proposed a class of diffusion LMS strategies, regularized by convex sparsifying penalties, for distributed estimation over adaptive networks.
Two different penalty functions have been employed: the $\ell_1$-norm, which uniformly attracts to zero all the vector elements, and a reweighted function, which better approximates the $\ell_0$-norm, selectively shrinking only the elements with small magnitude. Convergence and mean-square analysis of the sparse adaptive diffusion filter show under what conditions we have dominance of the proposed method with respect to its unregularized counterpart in terms of steady-state performance. Further analysis leads to a procedure to update the regularization parameter of the algorithm, in order to ensure dominance of the sparse diffusion filter with respect to its unregularized version.
In this way, the network can adjust in real-time the system parameters to improve the estimation
performance, according to the sparsity of the underlying vector. Several numerical results show the potential benefits of using such strategies.

\appendices

\section{Proof of Theorem 1}

Letting ${\cal B}=\mathcal{A}^T(I-\mathcal{M}\mathcal{D})$ and $b_i=\gamma\mathcal{A}^T\mathcal{M}\cdot\mathbb{E}\partial f(\bw_{i-1})$, recursion (\ref{compact_Diffusion_Expected}) gives
\begin{eqnarray}\label{recursion_expected}
\mathbb{E}\tilde{\bw}_i={\cal B}^i\mathbb{E}\tilde{\bw}_0+\sum_{n=0}^{i-1}{\cal B}^nb_{i-n}
\end{eqnarray}
where $\mathbb{E}\tilde{\bw}_0$ is the initial condition. As long as we can show that both terms on the right hand side of (\ref{recursion_expected}) converge as $i$ goes to infinity, then we would be able to conclude the convergence of $\mathbb{E}\tilde{\bw}_i$. To proceed, we call upon results from \cite{Takahashi,Chen-Sayed,Sayed2}.
Let $z=\mbox{\rm col}\{z_1,z_2,\ldots,z_N\}$ denote a vector that is obtained by stacking $N$ subvectors of size $M\times 1$ each (as is the case with $\tilde{\bw}_i$).  The block maximum norm of $z$ is defined as
\begin{equation}
\|z\|_{b,\infty}\;=\; \displaystyle \max_{1\leq k\leq N}\;\|z_k\|,
\end{equation}
where $\|\cdot\|$ denotes the Euclidean norm of its vector argument. Likewise, the induced block maximum norm of a block matrix ${\cal X}$ with $M\times M$ block entries is defined as:
\begin{equation}
\displaystyle \|{\cal X}\|_{b,\infty}\;=\ \max_{z\neq 0}\;\frac{\|{\cal X}z\|_{b,\infty}}{\|z\|_{b,\infty}}.
\end{equation}
It is easy to check that the first term on the RHS of (\ref{recursion_expected}) converges to zero as $i\rightarrow \infty$. Indeed, note that
\begin{equation}\label{conv_cond}
\|{\cal B}^i \mathbb{E}\tilde{\bw}_{0}\|_{\infty}\leq\|{\cal B}\|_{b,\infty}^i\cdot\|\mathbb{E}\tilde{\bw}_{0}\|_{b,\infty}\rightarrow 0
\end{equation}
if we can ensure that $\|{\cal B}\|_{b,\infty}<1$. This condition is actually satisfied by (\ref{step_sizes}). To see this, we invoke the triangle inequality of norms to note that
\begin{equation}\label{conv_cond}
\|{\cal B}\|_{b,\infty}=\left\|\mathcal{A}^T(I-\mathcal{M}\mathcal{D})\right\|_{b,\infty}\leq \left\|\mathcal{A}^T\right\|_{b,\infty}\cdot\|I-\mathcal{M}\mathcal{D}\|_{b,\infty}=\|I-\mathcal{M}\mathcal{D}\|_{b,\infty}
\end{equation}
since $\left\|\mathcal{A}^T\right\|_{b,\infty}=1$ in view of the fact that $A$ is a left-stochastic matrix \cite{Takahashi}. Therefore, to satisfy $\|{\cal B}\|_{b,\infty}<1$, it suffices to require
\begin{equation}\label{conv_cond2}
\|I-\mathcal{M}\mathcal{D}\|_{b,\infty}<1.
\end{equation}
Now, we recall a result from \cite{Chen-Sayed,Sayed} on the block maximum norm of a block diagonal and Hermitian matrix ${\cal X}$ with $M\times M$ blocks $\{X_k\}$, which states that
\begin{equation}\label{conv_cond3}
\|{\cal X}\|_{b,\infty}=\max_{k=1,\ldots,N}\rho(X_k)
\end{equation}
Thus, since ${\cal M}$ is diagonal, condition (\ref{conv_cond2}) will hold if the matrix $I-{\cal M}\mathcal{D}$ is stable. Using (\ref{perf_matrices3}), we can easily verify that this condition is satisfied for any step-sizes satisfying (\ref{step_sizes}), as claimed before. Therefore, when the step-sizes satisfy condition (\ref{step_sizes}), the first term on the RHS of (\ref{recursion_expected}) will converge to zero. We will show next that condition (\ref{step_sizes}) also implies that the second term on the RHS of (\ref{recursion_expected}) asymptotically converges to a finite value, thus leading to the overall convergence of the recursion (\ref{recursion_expected}).

One effective tool to prove convergence of a series is the comparison test \cite[p. 14]{Whittaker-Watson}: a series is absolutely convergent if each term of the series can be bounded by a term of an absolutely convergent series. Thus, denoting by $[x]_k$ the $k$-th entry of a vector $x$, it suffices to show that the series
\begin{eqnarray}\label{RHS2}
\sum_{n=0}^{\infty}\left[{\cal B}^nb_{i-n}\right]_k
\end{eqnarray}
converges for each $k=1,\ldots,NM$. Now, each term of the series in (\ref{RHS2}) can be bounded as:
\begin{eqnarray}\label{RHS2_bound}
\left[{\cal B}^nb_{i-n}\right]_k\leq\left|\left[{\cal B}^nb_{i-n}\right]_k\right|\leq\|{\cal B}^nb_{i-n}\|_{b,\infty}\leq\|{\cal B}\|^n_{b,\infty}\cdot\|b_{i-n}\|_{b,\infty}\leq \delta^n \cdot b_{\max}
\end{eqnarray}
where $\delta=\|{\cal B}\|_{b,\infty}$ and
\begin{eqnarray}
b_{\max}=\max_j\|\gamma\mathcal{A}^T\mathcal{M}\cdot\mathbb{E}\partial f(\bw_{j-1})\|_{b,\infty}
\end{eqnarray}
The second inequality in (\ref{RHS2_bound}) holds because the block maximum norm of a vector is greater than or equal to the largest absolute value of its entries. The scalar $b_{\max}$ is finite for the following reason. First, note that the subgradient vector $\partial f(\bw_{i-1})$ has bounded entries. In particular, $\partial f_{\max}\leq\sqrt{M}$ for the ZA update in (\ref{sign}), and $\partial f_{\max}\leq\sqrt{M}/\varepsilon$ for the RZA update in (\ref{update_rew_l1norm}).  We further note that $\left\|{\cal A}^T\right\|_{b,\infty}=1$ and $\displaystyle \|{\cal M}\|_{b,\infty}=\mu_{\max}$. It follows that
\begin{eqnarray}\label{bound_b}
b_{\max}\leq\max_i \gamma\cdot\|\mathcal{A}^T\|_{b,\infty}\cdot \|\mathcal{M}\|_{b,\infty} \cdot\|\mathbb{E}\partial f(\bw_{i-1})\|_{b,\infty} = \gamma \cdot \mu_{\max} \cdot \partial f_{\max}
\end{eqnarray}
Now, if condition (\ref{step_sizes}) is satisfied, then $\delta=\|{\cal B}\|_{b,\infty}<1$ and
\begin{eqnarray}\label{geom_series}
\sum_{n=0}^{\infty}\delta^n \cdot b_{\max}=\frac{b_{\max}}{1-\delta}
\end{eqnarray}
which means that the series (\ref{geom_series}) and, consequently, the series (\ref{RHS2}), are absolutely convergent. In summary, since both first and second term on the RHS of (\ref{recursion_expected}) asymptotically converge to finite values, we conclude that $\mathbb{E}\tilde{\bw}_i$ will converge to a steady-state value.
Now, taking the limit of (\ref{compact_Diffusion_Expected})  as $i\rightarrow\infty$, it is easy to derive a closed form expression for the bias:
\begin{eqnarray}\label{bias}
\hbox{bias}\triangleq\displaystyle\lim_{i\rightarrow\infty}\mathbb{E}\tilde{\bw}_i= \gamma\cdot\left[I-{\cal A}^T \left(I-{\cal M}\mathcal{D}\right)\right]^{-1}\displaystyle{\cal A}^T{\cal M}\lim_{i\rightarrow\infty} \mathbb{E} \partial f(\bw_{i-1})
\end{eqnarray}
Moreover, exploiting (\ref{RHS2_bound}), (\ref{bound_b}) and (\ref{geom_series}), we further note that
\begin{eqnarray}\label{recursion_expected3}
\|\hbox{bias}\|_{b,\infty}&\triangleq&\lim_{i\rightarrow\infty}\|\mathbb{E}\tilde{\bw}_i\|_{b,\infty}
=\lim_{i\rightarrow\infty}\left\|\sum_{n=0}^{i-1}{\cal B}^nb_{i-n}\right\|_{b,\infty}
\leq\lim_{i\rightarrow\infty} \sum_{n=0}^{i-1} \left\| {\cal B}^nb_{i-n}  \right\|_{b,\infty}\\
&\leq&\lim_{i\rightarrow\infty} \sum_{n=0}^{i-1} \left\| {\cal B}\right\|^n_{b,\infty} \left\| b_{i-n}\right\|_{b,\infty}
\leq \frac{b_{\max}}{1-\delta}\leq\frac{\gamma \cdot \mu_{\max} \cdot \partial f_{\max}}{1-\delta}
\end{eqnarray}
This completes the proof of Theorem 1.

\section{Proof of Theorem 2}

From (\ref{phiSigma})-(\ref{alpha}) we have
\begin{eqnarray}\label{phi_rho}
\phi_{\Sigma,i}(\gamma)= \gamma^2\mathbb{E}\|\partial f(\bw_{i-1})\|^2_{\mathcal{M}\mathcal{A}\Sigma\mathcal{A}^T\mathcal{M}}+2\gamma\mathbb{E}\partial f(\bw_{i-1})^T\mathcal{M}\mathcal{A}\Sigma\mathcal{A}^T\left(I-{\cal M}\mathcal{D}\right)\tilde{\bw}_{i-1}
\end{eqnarray}
Since, as noted in Appendix A, $\partial f(\cdot)$ is a bounded function for all $i$, the term $\beta_{\Sigma,i}$ in (\ref{beta}) can be upper bounded by a positive constant term $p_1$ for all $i$. The term $\alpha_{\Sigma,i}$ in (\ref{alpha}) can be written as $\mathbb{E}\bc_{i-1}^T\tilde{\bw}_{i-1}$ where the vector
\begin{eqnarray}\label{ci-1}
\bc_{i-1}\triangleq-2\left(I-{\cal M}\mathcal{D}\right)^T\mathcal{A}\Sigma\mathcal{A}^T\mathcal{M}\partial f(\bw_{i-1})
\end{eqnarray}
is again bounded for all $i$. Thus, we have
\begin{eqnarray}
\mathbb{E}\bc^T_{i-1}\tilde{\bw}_{i-1}
\leq\left|\sum_{m=1}^M\mathbb{E}\bc_{m,i-1}\tilde{\bw}_{m,i-1}\right|\leq c_{\rm max}\left|\sum_{m=1}^M\mathbb{E}\tilde{\bw}_{m,i-1}\right|\leq c_{\rm max} \mathds{1}^T\left|\mathbb{E}\tilde{\bw}_{i-1}\right|
\end{eqnarray}
where $c_{\rm max}=\max_i|\bc_{m,i-1}|$. As shown in Appendix A, the evolution of $\mathbb{E}\tilde{\bw}_{i-1}$ is given by (\ref{recursion_expected}), which, for any finite initialization vector $\tilde{w}_0$, converges as $i\rightarrow \infty$ and cannot diverge for all $i$, if the step-sizes are chosen to satisfy (\ref{step_sizes}). Consequently, $|\mathbb{E}\tilde{\bw}_{i-1}|$ can be upper bounded by some positive constant vector $p_2$ for all $i$.
Thus, letting $r={\rm vec}(\mathcal{A}^T\mathcal{M}{\cal G}^T\mathcal{M}\mathcal{A})$, expression (\ref{weighted_norm3}) can be upper bounded as
\begin{eqnarray}\label{MSDi2}
\mathbb{E}\|\tilde{\bw}_{i}\|^2_{\sigma}\leq\mathbb{E}\|\tilde{\bw}_{i-1}\|^2_{{\cal F}\sigma}+r^T\sigma+p_3
\end{eqnarray}
where $p_3=\gamma^2 p_1+\gamma c_{\rm max} \mathds{1}^Tp_2>0$. The positive constant $p_3$ can be related to the quantity $r^T\sigma$ through some  constant $\upsilon\in\mathbb{R}^+$, say, $p_3=\upsilon r^T\sigma$. Relation (\ref{MSDi2}) is an inequality, which can be used to prove convergence of the sequence $\mathbb{E}\|\tilde{\bw}_{i}\|^2_{\sigma}$ to a bounded region instead of a fixed point. Alternatively, we convert (\ref{MSDi2}) into an equality recursion as follows:
\begin{eqnarray}\label{MSDi3}
\mathbb{E}\|\tilde{\bw}_{i}\|^2_{\sigma}=\theta_{i}\mathbb{E}\|\tilde{\bw}_{i-1}\|^2_{{\cal F}\sigma}+\theta_{i}(1+\upsilon)r^T\sigma
\end{eqnarray}
for some coefficient $\theta_{i}\in[0,1]$ that depends on both $\mathbb{E}\|\tilde{\bw}_{i}\|^2_{\sigma}$ and $\mathbb{E}\|\tilde{\bw}_{i-1}\|^2_{{\cal F}\sigma}$. Recursion (\ref{MSDi3}) leads to:
\begin{eqnarray}\label{MSDi4}
\mathbb{E}\|\tilde{\bw}_{i}\|^2_{\sigma}=\left[ \prod_{l=1}^{i}\theta_l \right] \mathbb{E}\|\tilde{\bw}_{0}\|^2_{{\cal F}^i\sigma}+(1+\upsilon)r^T\sum_{l=0}^{i-1}\left[\prod_{n=i-l}^{i}\theta_n\right]{\cal F}^l\sigma
\end{eqnarray}
where $\mathbb{E}\|\tilde{\bw}_{0}\|^2$ is the initial condition.
We first note that if ${\cal F}$ is stable, ${\cal F}^i \rightarrow 0$ as $i\rightarrow\infty$. In this way, the first term on the RHS of (\ref{MSDi4}) vanishes asymptotically. Now, proceeding as in Appendix A, we can use the comparison test \cite[p. 14]{Whittaker-Watson} to prove that, if ${\cal F}$ is a stable matrix, the second term on the RHS of (\ref{MSDi4}) is an absolutely convergent series. Thus, denoting again by $[x]_k$ the $k$-th entry of a vector $x$, it suffices to show the convergence of the series:
\begin{eqnarray}\label{RHS3}
\sum_{l=0}^{\infty}\left[\xi_l(i){\cal F}^l\sigma\right]_k
\end{eqnarray}
with $\xi_l(i)=\prod_{n=i-l}^{i}\theta_n$, for $k=1,\ldots,NM$.  Each term of the series in (\ref{RHS3}) can be bounded as:
\begin{eqnarray}\label{RHS2_bound3}
\left[\xi_l(i){\cal F}^l\sigma\right]_k\leq\left|\left[\xi_l(i){\cal F}^l\sigma\right]_k\right|\leq\left|\left[{\cal F}^l\sigma\right]_k\right|\leq\|{\cal F}^l\sigma\|_{b,\infty}\leq\|{\cal F}^l\|_{b,\infty}\|\sigma\|_{b,\infty}
\end{eqnarray}
where the second inequality in (\ref{RHS2_bound3}) holds because the coefficients $\xi_l(i)\in[0,1]$ for all $i$, whereas the third inequality in (\ref{RHS2_bound3}) holds because the block maximum norm of a vector is greater equal than the largest absolute value of its entries.
A known result in matrix theory \cite[p. 30]{Kailath-Sayed} states that for every square stable matrix ${\cal F}$, and every $\epsilon>0$, there exists a submultiplicative matrix norm $\|\cdot\|_{\rho}$ such that
\begin{eqnarray}
\|{\cal F}\|_{\rho}=\rho({\cal F})+\epsilon
\end{eqnarray}
Since ${\cal F}$ is stable, $\rho({\cal F})<1$, we can choose $\epsilon>0$ such that $\rho({\cal F})+\epsilon=\xi<1$. Now, since in a finite dimensional space all norms are equivalent \cite{Horn-Johnson}, we have $\|\cdot\|_{b,\infty}\leq \zeta\cdot\|\cdot\|_{\rho}$, for some positive constant $\zeta$. Thus, we have
\begin{eqnarray}\label{boundsF}
\|{\cal F}^l\|_{b,\infty}\leq\zeta\cdot\|{\cal F}^l\|_{\rho}\leq\zeta\cdot\|{\cal F}\|^l_{\rho}=\zeta\cdot\xi^l
\end{eqnarray}
and, substituting (\ref{boundsF}) into (\ref{RHS2_bound3}), we get
\begin{eqnarray}\label{geom_series2}
\sum_{l=0}^{\infty} \|{\cal F}^l\|_{b,\infty}\cdot\|\sigma\|_{b,\infty}\leq \zeta\cdot\|\sigma\|_{b,\infty}\cdot  \sum_{l=0}^{\infty}\xi^l=\frac{\zeta\cdot\|\sigma\|_{b,\infty}}{1-\xi}
\end{eqnarray}
which means that the series (\ref{geom_series2}) and, consequently, the series (\ref{RHS3}), are absolutely convergent.
In summary, since both the first and second terms on the RHS of (\ref{MSDi4}) asymptotically converge to finite values, we conclude that $\mathbb{E}\|\tilde{\bw}_{i}\|^2_{\sigma}$ will converge to a steady-state value, thus completing our proof.

\section{Existence of $\alpha_{\Sigma,\infty}$}

Let us consider the bounded random vector $\bc_i$ in (\ref{ci-1}), which is independent of the noise sequence $\bv_k(i)$ for all $k,i$. Letting ${\cal B}=\mathcal{A}^T(I-\mathcal{M}\mathcal{D})$ and $\bb_i=\gamma\mathcal{A}^T\mathcal{M}\partial f(\bw_{i-1})$, from (\ref{compact_Diffusion}), we get
\begin{eqnarray}\label{recursion_expected_alpha}
\alpha_{\Sigma,\infty}=\lim_{i\rightarrow\infty}\mathbb{E}\bc_i^T\tilde{\bw}_i=\lim_{i\rightarrow\infty}\mathbb{E}\bc_i^T{\cal B}^i\tilde{\bw}_0+\lim_{i\rightarrow\infty}\sum_{n=0}^{i-1}\mathbb{E}\bc_i^T{\cal B}^n\bb_{i-n}
\end{eqnarray}
where $\tilde{\bw}_0$ is the initial condition. Following the same steps as in Appendix A, if the step-sizes satisfy condition (\ref{step_sizes}), the first term on the RHS of (\ref{recursion_expected_alpha}) will converge to zero. Furthermore, since the vector sequence $\bc_i$ is bounded, similarly to what we have done in (\ref{RHS2})-(\ref{geom_series}), we can again use the comparison test \cite[p. 14]{Whittaker-Watson} to prove that the second term on the RHS of (\ref{recursion_expected_alpha}) asymptotically converges to a finite value, thus leading to the existence of the limit in (\ref{recursion_expected_alpha}).

\begin{biography}[{\includegraphics[width=1in,height=1.25in,clip,keepaspectratio]{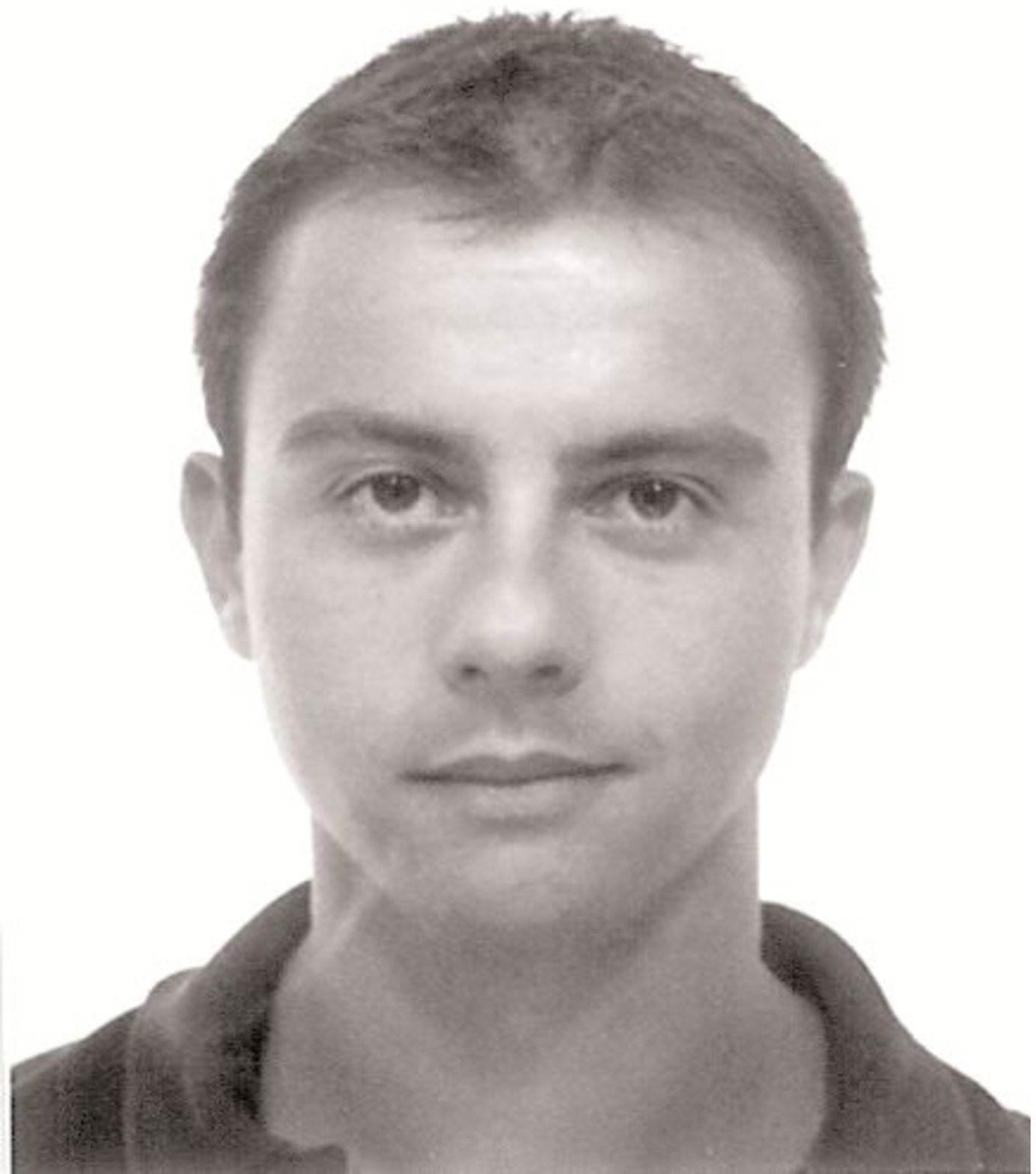}}]{Paolo Di Lorenzo}
(S'10)  received the M.Sc. degree in 2008 and the Ph.D. in electrical engineering in 2012, both from University of Rome
``La Sapienza,'' Italy. He is currently a post-doc in the Department of Information, Electronics and Telecommunications,
University of Rome, ``La Sapienza.'' During 2010 he held a visiting research appointment in the Department
of Electrical Engineering, University of California at Los Angeles (UCLA). He has participated in the European research project FREEDOM on femtocell networks. He is currently involved in the European projects SIMTISYS, on moving
target detection through satellite constellations, and TROPIC, on distributed computing, storage and radio resource allocation over cooperative femtocells. His primary research interests are in statistical signal processing, distributed
optimization algorithms for communication and sensor networks, graph theory, and adaptive filtering. Dr. Di Lorenzo received three best student paper awards, respectively at IEEE SPAWC'10, EURASIP EUSIPCO'11, and IEEE CAMSAP'11, for works in the area of signal processing for communications and synthetic aperture radar systems. He is recipient of the 2012 GTTI (Italian national group on telecommunications and information theory) award for the Best Ph.D. Thesis in information technologies and communications.
\end{biography}

\begin{biography}[{\includegraphics[width=1in,height=1.25in,clip,keepaspectratio]{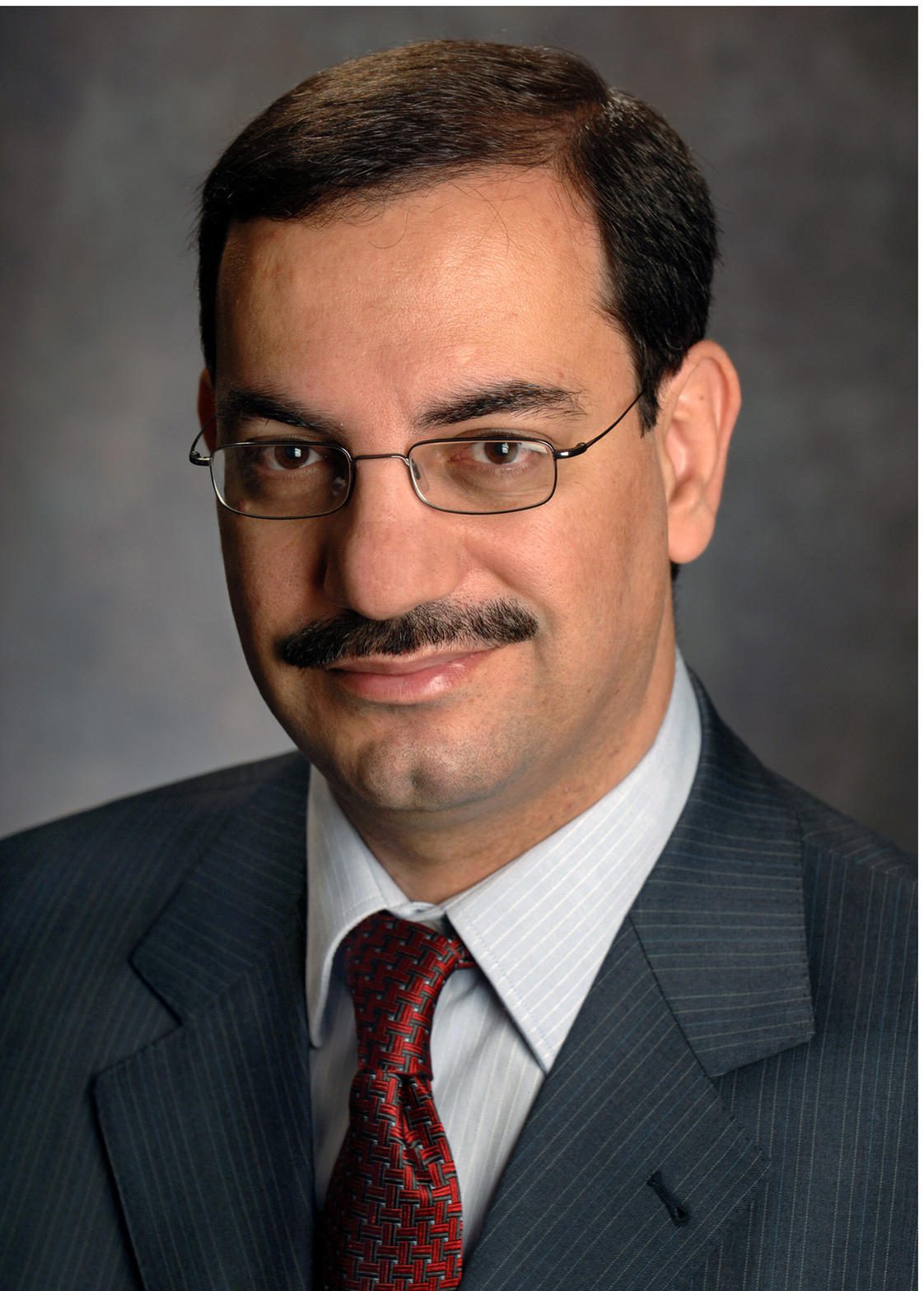}}]{Ali H. Sayed}
(S'90-M'92-SM'99-F'01) is Professor of Electrical Engineering
with the University of California, Los Angeles (UCLA), where he leads the
Adaptive Systems Laboratory. He has published widely, in the areas of
adaptation and learning, statistical signal processing, distributed
processing,  and bio-inspired cognition. He is coauthor of the
textbook \textit{Linear Estimation} (Englewood Cliffs, NJ: Prentice-Hall, 2000), of the research
monograph \textit{Indefinite Quadratic Estimation and Control} (Philadelphia, PA: SIAM, 1999), and
co-editor of \textit{Fast Algorithms for Matrices with Structure} (Philadelphia,
PA: SIAM, 1999). He is also the author of the textbooks \textit{Fundamentals of
Adaptive Filtering }(Hoboken, NJ: Wiley, 2003), and \textit{Adaptive Filters} (Hoboken, NJ: Wiley, 2008). He
has contributed several encyclopedia and handbook articles.
Dr. Sayed is a Fellow of IEEE for his contributions to adaptive ?ltering
and estimation algorithms. He has served on the Editorial Boards of several
publications. He has also served as the Editor-in-Chief of the IEEE
TRANSACTIONS ON SIGNAL PROCESSING from 2003 to 2005, and the EURASIP
Journal on Advances in Signal Processing from 2006 to 2007. He has served
on the Publications (2003-2005), Awards (2005), and Conference Boards
(2007-present) of the IEEE Signal Processing Society. He also served on the
Board of Governors of the IEEE Signal Processing Society from 2007 to 2008
and as Vice President of Publications of the same Society from 2009 to
2011. His work has received several recognitions, including the 1996 IEEE
Donald G. Fink Award, the 2002 Best Paper Award from the IEEE Signal
Processing Society, the 2003 Kuwait Prize in Basic Sciences, the 2005
Terman Award, and the 2005 Young Author
Best Paper Award from the IEEE Signal Processing Society. He has served
as a 2005 Distinguished Lecturer of the IEEE Signal Processing Society and
as General Chairman of the IEEE International Conference on Acoustics,
Speech, and Signal Processing (ICASSP) 2008.
\end{biography}

\end{document}